\documentclass[10pt,twocolumn,letterpaper]{article}

\usepackage[pagenumbers]{cvpr} 

\usepackage{xspace}
\usepackage{amssymb}
\usepackage{amsmath}
\usepackage{amsfonts}
\usepackage{wrapfig}
\usepackage{algorithm}
\usepackage{algorithmic}

\usepackage{multicol}
\usepackage{multirow}
\usepackage{makecell}
\usepackage{wrapfig}
\usepackage{tabularx}
\usepackage{adjustbox}

\usepackage{pifont}
\usepackage{color}
\usepackage{colortbl}
\usepackage[misc]{ifsym}
\usepackage{soul}
\usepackage{bm}
\usepackage{wrapfig}

\definecolor{lightgray}{rgb}{0.8, 0.8, 0.8}
\definecolor{lgray}{rgb}{0.66, 0.66, 0.66}

\definecolor{lblu_tab}{RGB}{225, 235, 246}
\definecolor{orange_vitad}{RGB}{222, 131, 68}
\definecolor{blue_vitad}{RGB}{106, 153, 208}
\definecolor{trajectory_green}{RGB}{126, 171, 85}
\definecolor{trajectory_yellow}{RGB}{245, 194, 66}
\definecolor{tab_others}{RGB}{235, 235, 235}
\definecolor{tab_ours}{RGB}{225, 235, 246}

\definecolor{whit_tab}{RGB}{255, 255, 255}
\definecolor{gray_tab}{RGB}{246, 246, 246}
\definecolor{oran_tab}{RGB}{252, 242, 237}
\definecolor{blue_tab}{RGB}{227, 240, 251}

\definecolor{cvprblue}{rgb}{0.21,0.49,0.74}
\usepackage[pagebackref,breaklinks,colorlinks,allcolors=cvprblue]{hyperref}

\def\paperID{3608} 
\def\confName{CVPR}
\def\confYear{2025}

\title{MobileMamba: Lightweight Multi-Receptive Visual Mamba Network}

\author{%
  Haoyang He$^{1*}$
  ~~ Jiangning Zhang$^{1,2}$\thanks{Equal contributions.}
  ~~ Yuxuan Cai$^3$
  ~~ Hongxu Chen$^1$
  ~~ Xiaobin Hu$^2$\\
  ~~ Zhenye Gan$^2$
  ~~ Yabiao Wang$^2$
  ~~ Chengjie Wang$^2$
  ~~ Yunsheng Wu$^2$
  ~~ Lei Xie$^1$\thanks{Corresponding author.} \\
  \normalsize $^1$Zhejiang University ~~ $^2$Youtu Lab, Tencent ~~ $^3$Huazhong University of Science and Technology\\
  \normalsize Code: \textcolor{red}{\url{https://github.com/lewandofskee/MobileMamba}}
}

\begin{document}
\maketitle
\begin{abstract}
Previous research on lightweight models has primarily focused on CNNs and Transformer-based designs. CNNs, with their local receptive fields, struggle to capture long-range dependencies, while Transformers, despite their global modeling capabilities, are limited by quadratic computational complexity in high-resolution scenarios. Recently, state-space models have gained popularity in the visual domain due to their linear computational complexity. Despite their low FLOPs, current lightweight Mamba-based models exhibit suboptimal throughput.
In this work, we propose the \textbf{MobileMamba} framework, which balances efficiency and performance. We design a three-stage network to enhance inference speed significantly. At a fine-grained level, we introduce the \textbf{M}ulti-\textbf{R}eceptive \textbf{F}ield \textbf{F}eature \textbf{I}nteraction~(MRFFI) module, comprising the Long-Range \textbf{W}avelet \textbf{T}ransform-\textbf{E}nhanced \textbf{Mamba}~(WTE-Mamba), Efficient \textbf{M}ulti-\textbf{K}ernel \textbf{De}pthwise \textbf{Conv}olution~(MK-DeConv), and Eliminate Redundant \textbf{Identity} components. This module integrates multi-receptive field information and enhances high-frequency detail extraction. Additionally, we employ training and testing strategies to further improve performance and efficiency.
MobileMamba achieves up to \textbf{83.6\%} on Top-1, surpassing existing state-of-the-art methods which is maximum \textbf{$\times$21$\uparrow$} faster than LocalVim on GPU. Extensive experiments on high-resolution downstream tasks demonstrate that MobileMamba surpasses current efficient models, achieving an optimal balance between speed and accuracy.
The full code is available in
\url{https://github.com/lewandofskee/MobileMamba}.

\end{abstract}    
\vspace{-0.48cm}
\section{Introduction} \label{sec:introduction}
\begin{figure}[tp]
\centering
\includegraphics[width=0.48\textwidth]{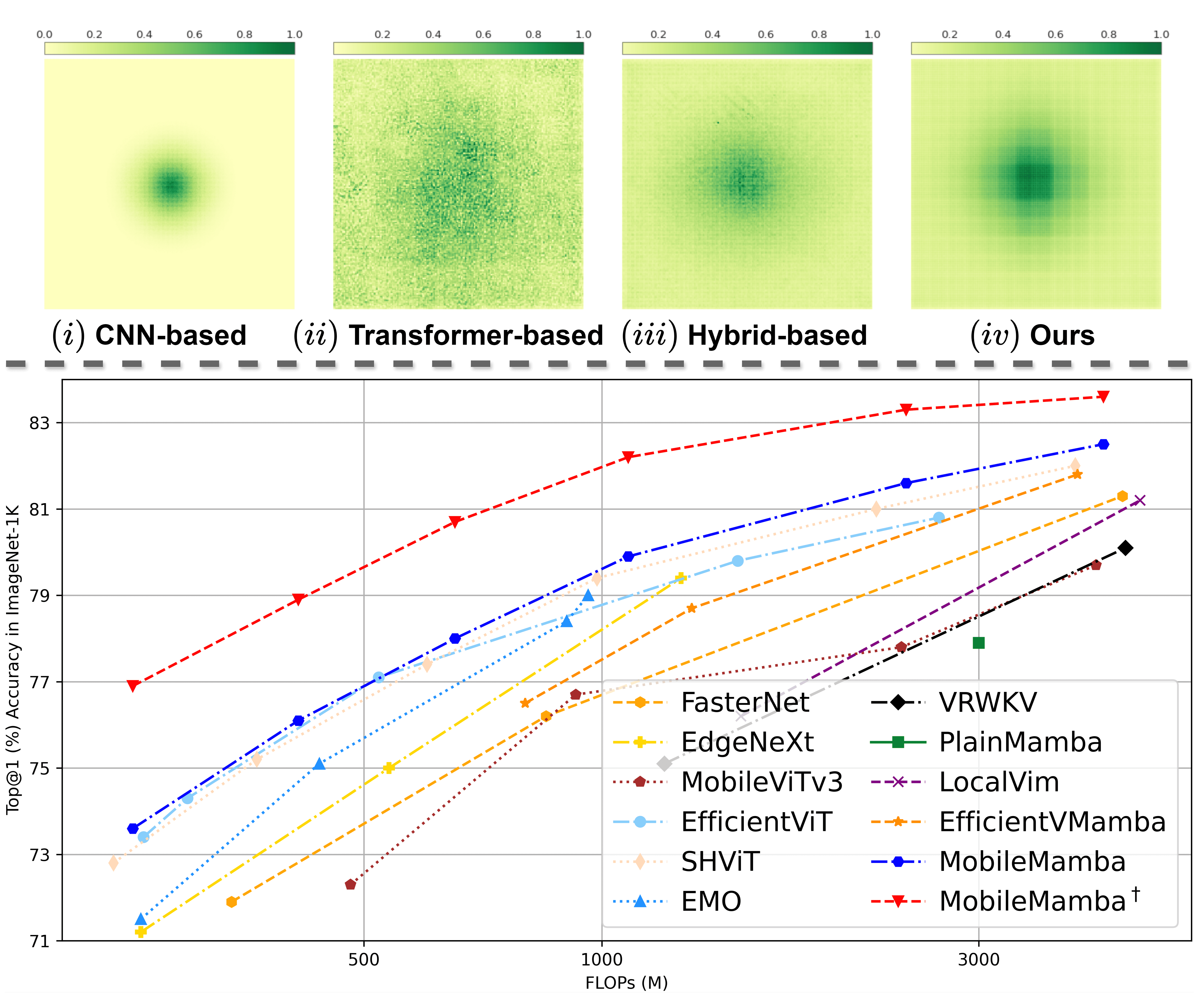}
\caption{\textbf{Top}: Visualization of the \textit{Effective Receptive Fields}~(ERF) for different architectures. \textbf{Bottom}: Performance \textit{vs.} FLOPs with recent CNN/Transformer/Mamba-based methods.}
\label{fig:motiv}
\vspace{-0.48cm}
\end{figure}

The proliferation of mobile devices has increased the demand for efficient and accurate visual processing in resource-constrained environments. Lightweight models significantly reduce computational and storage costs while enhancing inference speed. Current lightweight models are primarily categorized into CNN-based and Transformer-based structures. CNN-based MobileNets~\cite{mobilenet,mobilenetv2,mobilenetv3} use depth-wise separable convolutions to reduce computational complexity, laying a foundation for subsequent CNN-based work~\cite{ghostnetv2,fasternet,edgenext,efficientnet,shufflenet}. However, the major drawback of CNN-based methods is their local \textbf{E}ffective \textbf{R}eceptive \textbf{F}ield (ERF), as shown in Fig.~\ref{fig:motiv}\textit{(\romannumeral1)}, which is \textit{confined to the central region and lacks long-range correlations.} In downstream tasks~(Tab.~\ref{table:det_coco_ssdlite}) with high-resolution inputs, CNN-based methods can only achieve performance improvements by increasing computational load.

\textbf{Vi}sion \textbf{T}ransformers (ViTs) exhibit a global ERF and long-range modeling capabilities in Fig.~\ref{fig:motiv}\textit{(\romannumeral2)}. However, their quadratic computational complexity results in higher overhead compared to CNNs. Some works~\cite{efficientformer,efficientformerv2,efficientvit,shvit,mobilevit,mobilenetv2,mobilevitv3} have reduced resolution or channel count to alleviate this complexity, achieving notable results. Despite this, pure ViTs lack inductive bias, prompting researchers to develop hybrid CNN-ViT structures~\cite{emo,metaformer,mpvit} that combine local and global ERF for improved performance in Fig.~\ref{fig:motiv}\textit{(\romannumeral3)}. However, \textit{ViT-based methods still face the issue of quadratic computational complexity, especially with high-resolution inputs in downstream tasks~(Tab.~\ref{table:seg_three}).}

\begin{wrapfigure}{r}{0.32\textwidth} \vspace{-1em}
\includegraphics[width=0.32\textwidth]{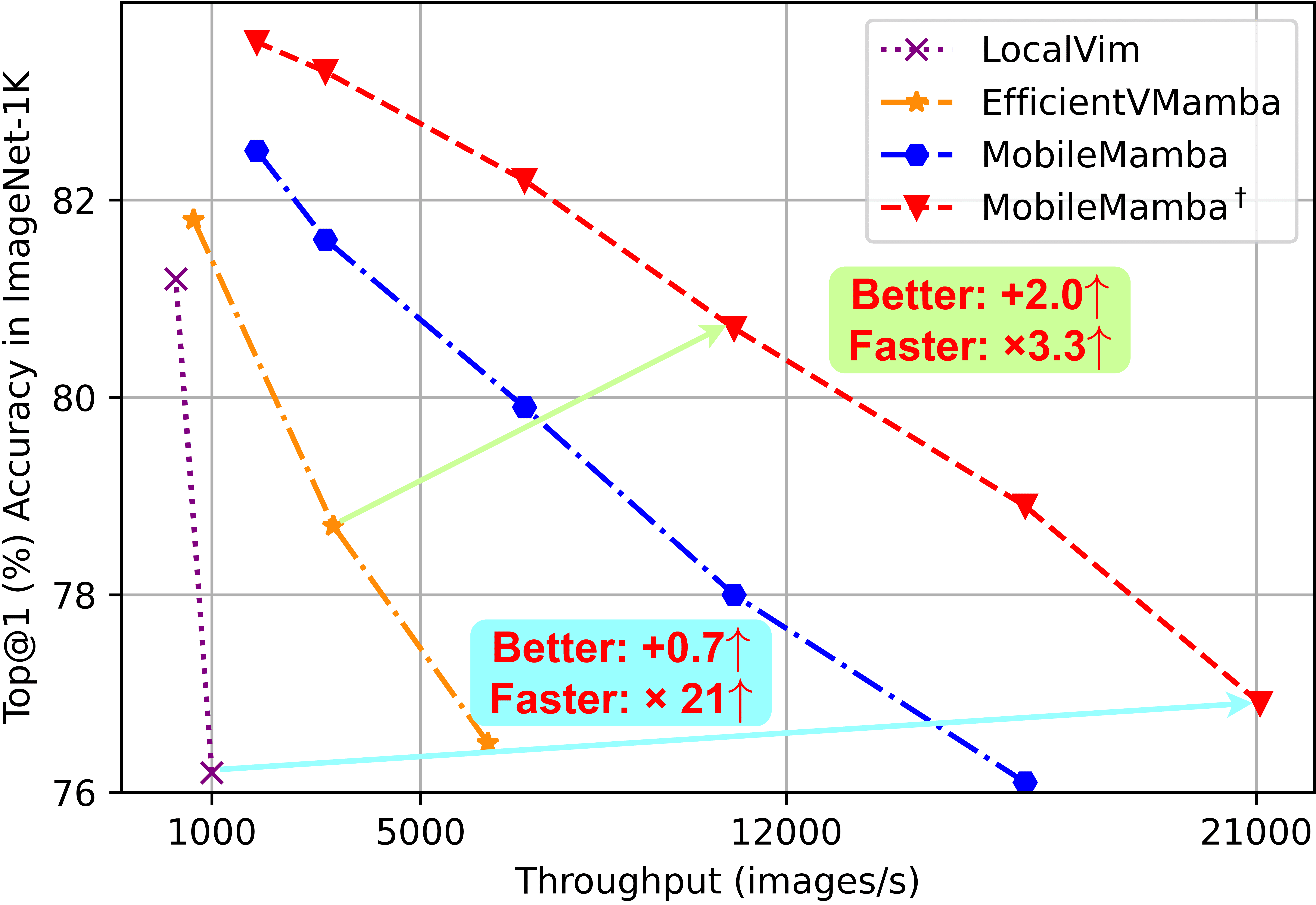}
\caption{\textbf{Accuracy \textit{vs.} Speed} with Mamba-based methods.}
\label{fig:speedmamba}
\vspace{-2em}
\end{wrapfigure}

State-space models~\cite{s4,s5,h3,mamba} have gained attention for capturing long-range dependencies with linear computational complexity. Researchers have successfully applied these models to the visual domain~\cite{vim,vmamba,msvmamba}, achieving notable effectiveness and efficiency. 
The recent lightweight Mamba-based models~\cite{localmamba,efficientvmamba} introduce different efficient scanning methods to reduce complexity.
However, only FLOPs are reported in their works, which do not necessarily correlate with fast inference speed. Experimental results in Fig.~\ref{fig:speedmamba} show that current Mamba-based structures suffer from slow inference speeds and poor performance.

Based on the above motivation, we propose \textbf{MobileMamba}, designed as an efficient lightweight network through \textit{Coarse-Grained}, \textit{Fine-Grained}, and \textit{Training/Testing Strategies}. Firstly, in Sec.~\ref{coarse}, we discuss the trade-offs between four-stage and three-stage networks in terms of accuracy, speed, and FLOPs. As shown in Fig.~\ref{fig:macro}, under the same throughput, a three-stage network achieves higher accuracy. Similarly, for the same performance, a three-stage network has higher throughput. Therefore, we select a three-stage network as our Coarse-Grained framework.
In the design of the MobileMamba module in Sec.~\ref{finegrained}, we introduce an efficient \textbf{M}ulti-\textbf{R}eceptive \textbf{F}ield \textbf{F}eature \textbf{I}nteraction (MRFFI) module. Specifically, the input features are divided into three parts along the channel dimension. The first part uses a Long-Range \textit{\textbf{W}avelet \textbf{T}ransform-\textbf{E}nhanced Mamba}~(WTE-Mamba) module to extract global features while enhancing the extraction of fine-grained details such as edge information. The second part employs \textit{\textbf{M}ulti-\textbf{K}ernel \textbf{De}pthwise Convolution}~(MK-DeConv) operations to capture multi-scale receptive fields. The final part uses \textit{Eliminate redundant} \textit{\textbf{Identity}} mapping to reduce channel redundancy in high-dimensional space, decreasing computational complexity and increasing processing speed. The features obtained through \textit{MRFFI integrate global and multi-scale local receptive field information, enhancing the extraction of high-frequency edge details.}
Finally, we enhance the model's learning capability through two training phase strategies in Sec.~\ref{strategies}, Knowledge Distillation, and Extended Training Epochs. Additionally, a Normalization Layer Fusion strategy in the testing phase improves the model's inference speed. 

In Fig.~\ref{fig:motiv}\textit{(iv)}, our approach utilizes a global ERF, whereas multi-kernel local convolution operations facilitate the extraction of adjacent information. The comparison with SoTA methods at the bottom of Fig.~\ref{fig:motiv} shows that MobileMamba$^\dagger$ (with training strategies) achieves Top-1 accuracies of 76.9/78.9/80.7/82.2/83.3/83.6 on ImageNet-1K~\cite{imagenet} for models ranging from 200M to 4G FLOPs, surpassing existing CNN, ViT, and Mamba-based methods. Compared to efficient Mamba-based methods in Fig.~\ref{fig:speedmamba}, 
MobileMamba improves Top-1 by +0.7$\uparrow$ while being \textbf{$\times$21$\uparrow$} times faster than LocalVim~\cite{localvim}, and improves by +2.0$\uparrow$ while being \textbf{$\times$3.3$\uparrow$} times faster than EfficientVMamba~\cite{efficientvmamba}.
This demonstrates a significant advantage over existing Mamba-based lightweight model designs.
Extensive experiments on downstream tasks further validate the effectiveness of our method. On Mask RCNN~\cite{maskrcnn}, MobileMamba improves $mAP^b$ by +1.3$\uparrow$, $mAP^m$ by +1.0$\uparrow$ and throughput by +56\%$\uparrow$ compared to EMO~\cite{emo}. On RetinaNet~\cite{retinanet}, it improves $mAP^b$ by +2.1$\uparrow$ and throughput by \textbf{$\times$4.3$\uparrow$} compared to EfficientVMamba~\cite{efficientvmamba}. On SSDLite~\cite{ssdlite}, it achieves mAPs of 24.0/29.5 by increasing resolution. On DeepLabv3~\cite{deeplabv3}, Semantic FPN~\cite{segfpn}, and PSPNet~\cite{pspnet}, it achieves mIoUs of up to 37.4/42.5/36.9 with fewer FLOPs. Compared to CNN-based MobileNetv2~\cite{mobilenetv2} and ViT-based MobileViTv2~\cite{mobilevitv2}, our approach achieves improvements of +7.2$\uparrow$ and +0.4$\uparrow$, respectively, in high-resolution 512x512 input downstream tasks, while only requiring \textbf{8.5\%} and \textbf{11.2\%} of their FLOPs for PSPNet~\cite{pspnet}.

In summary, our contributions are as follows:
\begin{itemize}
  \item We propose a lightweight three-stage MobileMamba framework that achieves a good balance between performance and efficiency. The effectiveness and efficiency of MobileMamba have been validated on classification tasks as well as three high-resolution input downstream tasks.
  \item We designed an efficient Multi-Receptive Field Feature Interaction (MRFFI) module to enhance multi-scale perception capabilities with larger ERF and improve the extraction of fine-grained high-frequency edge information. 
  \item MobileMamba significantly enhances performance and efficiency by employing training and testing strategies across a range of models of different FLOPs sizes. 
\end{itemize}
\section{Related Work} \label{sec:related_work}
\subsection{Lightweight Visual Models}
The most extensively studied lightweight visual networks can be categorized into CNN-based and Vision Transformer (ViT)-based structures. 
The CNN-based MobileNets~\cite{mobilenet,mobilenetv2,mobilenetv3} transitions from standard convolution to depthwise separable convolution, significantly reducing computational complexity.
GhostNets~\cite{ghostnet,ghostnetv2,ghostnetv3} replaces the original convolution with a cheap operation on half of the channels.
Additionally, numerous CNN-based works~\cite{shufflenet,shufflenetv2,efficientnet,efficientnetv2,fasternet}demonstrate excellent performance and efficiency on mobile devices.
The main limitation of these methods is their local receptive field. In contrast, ViT possesses a global receptive field and the ability to capture long-range dependencies. Nevertheless, their quadratic computational complexity results in higher computational costs compared to CNNs. Therefore, lightweight vision Transformers are designed to retain their global receptive field while reducing computational overhead.
EfficientViT~\cite{efficientvit} designs a three-stage network and proposes Cascaded Group Attention to significantly improve inference speed.
SHViT~\cite{shvit} introduces Single Head Self-Attention, selecting only a few channels to use ViT while directly connecting the remaining channels via Identity, greatly enhancing operational efficiency.
Furthermore, many hybrid methods~\cite{fastvit,edgevits,emo,efficientformer,efficientformerv2,mobilevit,mobilevitv2,mobilevitv3}, have achieved outstanding performance.

\subsection{State Space Models}
\noindent State Space Models (SSMs)~\cite{s4,s5,ssm2,ssm3,h3} inspired by control systems, can be regarded as  linear time-invariant systems mapping input $x(t) \in \mathbb{R}^L$ to output $y(t) \in \mathbb{R}^L$ via hidden state $h(t) \in \mathbb{R}^M$: $h^{\prime}(t) = \mathbf{A} h(t) + \mathbf{B} x(t), y(t) = \mathbf{C} h(t),$ where $\mathbf{A} \in \mathbb{R}^{M \times M}$, $\mathbf{B} \in \mathbb{R}^{M \times 1}$, and $\mathbf{C} \in \mathbb{R}^{1 \times M}$.

Mamba~\cite{mamba} uses zero-order hold with timescale $\Delta$ to convert continuous $\mathbf{A}$ and $\mathbf{B}$ to discrete $\overline{\mathbf{A}}$ and $\overline{\mathbf{B}}$:
\vspace{-0.3em}
\begin{equation}
    \begin{aligned}
        \overline{\mathbf{A}} &= \exp(\Delta \mathbf{A}), ~~~
        \overline{\mathbf{B}} = (\Delta \mathbf{A})^{-1}(\exp(\Delta \mathbf{A}) - \mathbf{I}) \cdot \Delta \mathbf{B}.
    \end{aligned}
    \vspace{-0.3em}
\end{equation}
The discrete system is: $ h_t = \overline{\mathbf{A}} h_{t-1} + \overline{\mathbf{B}} x_t, y_t = \mathbf{C} h_t.$
From a global convolution perspective:
\vspace{-0.3em}
\begin{equation}
\begin{aligned}
\overline{\mathbf{K}} &= (\mathbf{C} \overline{\mathbf{B}}, \mathbf{C} \overline{\mathbf{A}} \overline{\mathbf{B}}, \dots, \mathbf{C} \overline{\mathbf{A}}^{L-1} \overline{\mathbf{B}}), ~~~
\mathbf{y} = \mathbf{x} * \overline{\mathbf{K}},
\end{aligned}
\vspace{-0.3em}
\end{equation}
where $*$ is convolution, $L$ is the sequence length, and $\overline{\mathbf{K}} \in \mathbb{R}^L$ is the SSM kernel.

\noindent\textbf{SSMs for Vision.} SSMs~\cite{s4,s5,h3} have garnered significant attention due to their efficient computational complexity in capturing long-range dependencies. 
Mamba~\cite{mamba} introduces the S6 module,
achieving a simple structure with excellent efficiency in long-sequence modeling. Due to this advantage, numerous works have applied it to visual tasks~\cite{vim,vmamba,vmunet,umamba,plainmamba,mambavision,mambaad}. Vim~\cite{vim} proposes a bidirectional Mamba block, demonstrating its speed and memory advantages over ViTs at high resolutions. VMamba~\cite{vmamba} introduces Cross-Scan to enhance modeling capabilities. EfficientVMamba~\cite{efficientvmamba} proposes Efficient Scan, improving scanning efficiency through skip sampling.
LocalVim~\cite{localmamba} proposes local window scanning to enhance local information acquisition.
Despite various designs, no lightweight Mamba-based network surpasses existing CNN and ViT methods. This paper explores lightweight Mamba-based visual networks to achieve better performance, lower computational complexity, and faster inference speed.
\section{Methdology} \label{sec:method}
\subsection{Coarse-Grained Design of MobileMamba}
\label{coarse}
\begin{figure}[t]
\centering
\includegraphics[width=0.48\textwidth]{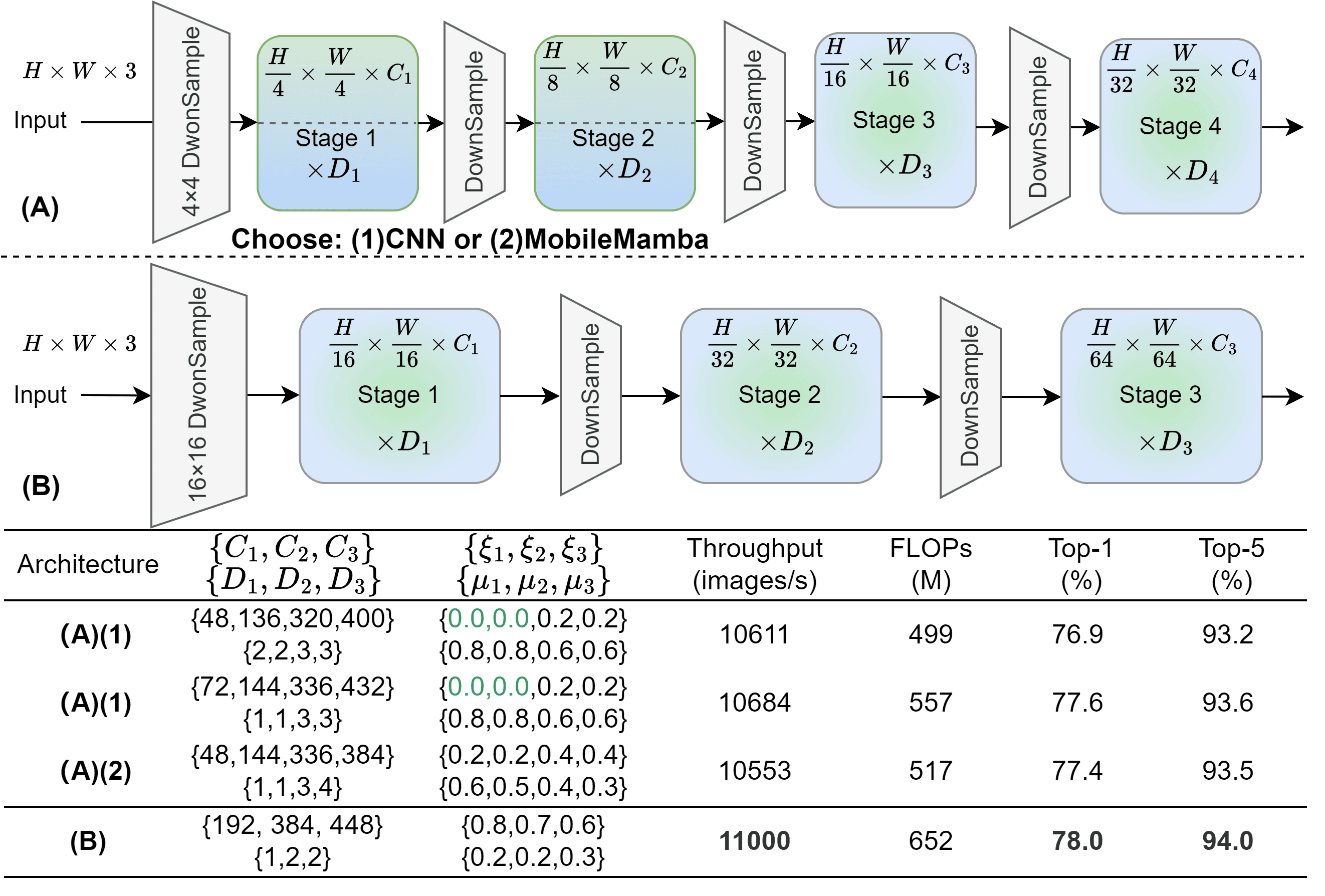} 
\caption{\textbf{Coarse-Grained Design}. (A) illustrates the structure of a commonly used four-stage network, where the first two stages can be configured with either (1) a purely CNN-based structure or (2) the MobileMamba structure. (B) depicts the three-stage network structure employed in this study. The following table presents the model parameters for different structures and the ImageNet-1K Top-1 and Top-5 at equivalent throughput.}
\label{fig:macro}
 \vspace{-0.48cm}
\end{figure}

In this section, we design the efficient MobileMamba structure, which includes a three-stage network as shown in Fig.~\ref{fig:macro}(B). Most existing network~\cite{metaformer,efficientformer,fasternet} follow the four-stage framework depicted in Fig.~\ref{fig:macro}(A). Specifically, in a four-stage network, the first downsampling reduces the input image $H\times W\times 3$ to $\frac{H}{4}\times \frac{W}{4}\times C_1$, and the final output feature map is $\frac{H}{32}\times \frac{W}{32}\times C_4$. In contrast, the three-stage network reduces the input image to $\frac{H}{16}\times \frac{W}{16}\times C_1$ during the first downsampling, and the final output feature map is $\frac{H}{64}\times \frac{W}{64}\times C_4$. Due to the larger feature map size in the four-stage network, it requires more computation and consequently operates at a slower speed.
The table below Fig.~\ref{fig:macro} compares the classification results on the ImageNet-1K~\cite{imagenet} dataset for the three-stage network and various four-stage networks under similar throughput conditions. In the first two experiments, the initial two stages of the four-stage network are designed with a purely CNN architecture, which enhances inference speed. The third experiment employs the MobileMamba blocks across all four stages of the network. The results indicate that although the four-stage network with a purely CNN structure in the first two stages shows improved inference speed and performance,\textit{ the three-stage network achieves faster inference with both Top-1 and Top-5 accuracy improvement of +0.4$\uparrow$.} Ultimately, we select the three-stage network structure to enhance inference speed and improve classification results.

\begin{figure*}[t]
\centering
\includegraphics[width=0.85\textwidth]{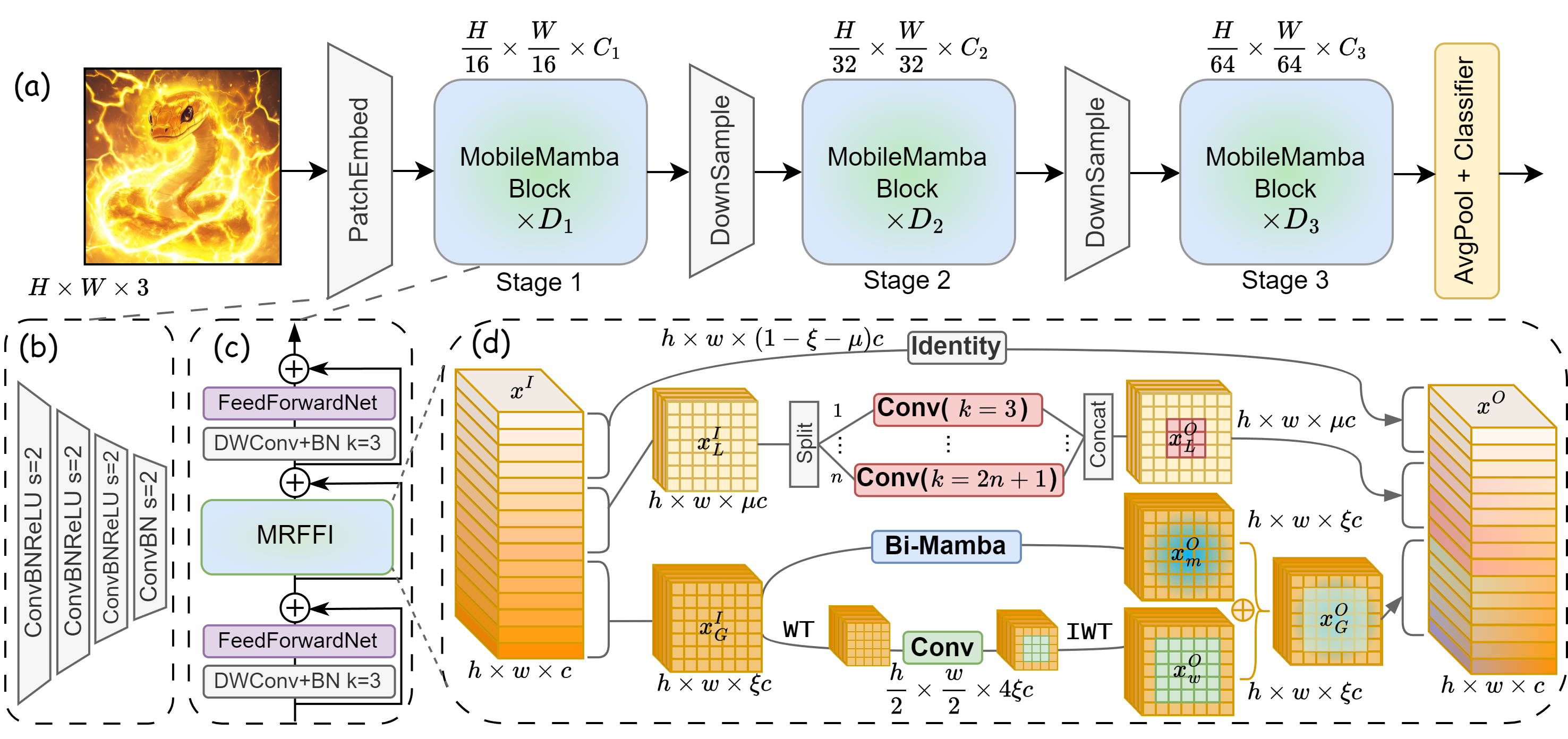} 
\caption{\textbf{Overview of MobileMamba}. (a) Architecture of MobileMamba. (b) 16 $\times$16 DownSample PatchEmbed. (c) Structure of MobileMamba Block. (d) \textbf{Fine-Grained Design.} The proposed efficient Multi-Receptive Field Feature Interaction~(MRFFI) module.}
\label{fig:mobilemamba}
\vspace{-1em}
\end{figure*}
\subsection{Fine-Grained Design of MobileMamba}
\label{finegrained}
The proposed efficient \textbf{M}ulti-\textbf{R}eceptive \textbf{F}ield \textbf{F}eature \textbf{I}nteraction~(\textbf{MRFFI}) module is placed between symmetric local information perception and an FFN in each MobileMamba block.
In the MRFFI module, features are divided into three parts along the channel dimension. \textit{1)}. The first part of the features undergoes the Long-Range \textit{\textbf{W}avelet \textbf{T}ransform-\textbf{E}nhanced \textbf{Mamba} (WTE-Mamba)}, which enhances the extraction of high-frequency edge information while performing global modeling. \textit{2)}. The second part is processed through \textit{\textbf{M}ulti-\textbf{K}ernel \textbf{De}pthwise \textbf{Conv}olution (MK-DeConv)} operations to enhance the perception capability of different receptive fields. \textit{3)}. The remaining features are subjected to \textit{Identity} mapping to reduce feature redundancy in high-dimensional space and decrease computational complexity, thereby improving processing speed.

\noindent\textbf{Long-range WTE-Mamba}. The purpose is to enhance the ability to extract \textit{fine-grained information}, such as \textit{high-frequency edge details, based on global modeling}. Also, the convolution operations~\cite{wct} on the WT feature maps have a \textit{larger ERF} compared to normal scales and exhibit \textit{lower computational complexity}. For the input feature $x^I \in \mathbb{R}^{h\times w \times c}$, the features $x^I_G \in \mathbb{R}^{h\times w \times \xi c}$ are processed through a bidirectional scanning Mamba module to learn global information, with the global channel proportion denoted as $0\leq\xi\leq1$. 
\vspace{-0.3em}
\begin{equation}
\begin{aligned}
& x^I_{m1}=\operatorname{SSM}\left(\sigma\left(\operatorname{Conv}\left(\operatorname{Linear}\left(x^I_G\right)[:\xi c]\right)\right)\right), \\
& x^I_{m2}=\sigma\left(\operatorname{Linear}\left(x^I_G\right)[\xi c:]\right),\\
& x^O_{m}=\operatorname{Linear}\left(x^I_{m1}\otimes x^I_{m2}\right).
\end{aligned}
\end{equation}

\vspace{-0.3em}
Simultaneously, the same feature map undergoes Haar Wavelet transformation to obtain feature representations $x^I_w \in \mathbb{R}^{\frac{h}{2}\times \frac{w}{2} \times 4\xi c}$ at different frequency scales. Local convolutional information extraction and Inverse Wavelet Transformation~(IWT) are then performed to restore the original feature map size $x^O_w \in \mathbb{R}^{h\times w \times \xi c}$. 
\vspace{-0.3em}
\begin{equation}
\begin{aligned}
& 
x^I_{wt} =\operatorname{WT}\left(x^I_w, \left[f_{L L}, f_{L H}, f_{H L}, f_{H H}\right]\right), \\
& x^O_{w} =\operatorname{IWT}\left(\operatorname{Conv}(x^I_{wt}), \left[f_{L L}, f_{L H}, f_{H L}, f_{H H}\right]\right),
\end{aligned}
\end{equation}
\renewcommand{\arraystretch}{0.8}
\setlength{\arraycolsep}{2pt}
\vspace{-0.3em}
where $f_{L L}=\frac{1}{2}\begin{bmatrix}1 & 1 \\ 1 & 1 \end{bmatrix}$ is a low-pass filter, and $f_{L H}=\frac{1}{2}\begin{bmatrix}
1 & -1 \\
1 & -1
\end{bmatrix}$, $f_{H L}=\frac{1}{2}\begin{bmatrix}
1 & 1 \\
-1 & -1
\end{bmatrix}$, $f_{H H}=\frac{1}{2}\begin{bmatrix}
1 & -1 \\
-1 & 1
\end{bmatrix}$ are a set of high filters. The final output feature map for this part is obtained by adding the output feature map from the Mamba module, which has extracted global information, to the output feature map from the wavelet-transformed and convolved local information.
\vspace{-0.3em}
\begin{equation}
x^O_G = x^O_{m} + x^O_{w}, \text{where}~x^O_G \in \mathbb{R}^{h\times w \times \xi c},
\end{equation}

\vspace{-0.3em}
\noindent\textbf{Efficient MK-DeConv}. This approach extracts local information with \textit{varying ERF}, enabling \textit{multi-receptive field interaction.} For the remaining features, $x^I_L \in \mathbb{R}^{h\times w \times \mu c}$ are selected, where the local channel proportion is denoted as $\mu\leq1-\xi$. These channels are then divided into $n \in \mathbb{N}$ parts. Each part $x^I_{Lj} \in \mathbb{R}^{h\times w \times \frac{\mu c}{n}}$ undergoes local convolution operations with different kernel sizes. Finally, the results from the different convolution operations are concatenated to form the output features $x^O_L \in \mathbb{R}^{h\times w \times \mu c}$. 
\vspace{-0.3em}
\begin{equation}
\begin{aligned}
& 
x^O_{Lj} =\operatorname{Conv}\left(x^I_{Lj}, \text{k}=(2j+1)\right), j\in 1,...,n.\\
& x^O_L =\operatorname{Concat}([x^O_{L1}, ..., x^O_{Ln}], \text{dim}=-1), \\
\end{aligned}
\end{equation}

\vspace{-0.3em}
\noindent \textbf{Eliminate redundant Identity}. Finally, to \textit{reduce the issue of feature redundancy in high-dimensional space}~\cite{ghostnet}, we apply identity mapping to the remaining $(1-\xi-\mu)c$ channels. This approach minimizes unnecessary computations and enhances operational efficiency. Therefore, the final output after processing through the MRFFI module is computed as follows:
\vspace{-0.3em}
\begin{equation}
x^O = \operatorname{Concat}(x^O_G, x^O_L, x^I[(1-\xi-\mu)c:]).
\end{equation}

\vspace{-0.3em}
MobileMamba is designed with six structures in Tab.~\ref{tab:architecture}. Across different models, we maintain the same global and local channel proportions. For the small model, we use a \textit{smaller input resolution to achieve lower computational complexity and faster runtime}. Conversely, for the large model, we use a \textit{larger input resolution to obtain better performance}, as detailed in Sec.~\ref{ablation}. By adjusting the input resolution according to the model size, we balance the trade-offs between computational efficiency and performance. This design strategy ensures that MobileMamba can be effectively scaled to meet different requirements while maintaining consistent channel proportions.

\begin{wraptable}{r}{0.28\textwidth}
  \centering
  \vspace{-0.26cm}
  \renewcommand{\arraystretch}{1.1}
    \setlength\tabcolsep{5.0pt}
  \resizebox{1\linewidth}{!}{
    \begin{tabular}{cccccc}
        \toprule[0.1em]
        Method & FLOPs & Throughput & Params & Top-1 \\
    \midrule
    TTT~\cite{ttt}   & 625   & 9569  & 14.2  & 77.0  \\
    xLSTM~\cite{xlstm} & 695   & 6868  & 14.6  & 77.3  \\
    RWKV6~\cite{rwkv6} & 658   & 10331  & 14.8  & 77.8 \\
    Mamba~\cite{mamba} & 652   & \textbf{11000} & 15.0  & \textbf{78.0} \\
        \toprule[0.1em]
        \end{tabular}
    }
    \vspace{-0.2cm}
    \caption{Other RNN Paradigm models.}
    \label{tab:rnn}%
    \vspace{-0.48cm}
\end{wraptable}
\noindent\textbf{Replace Mamba with other RNN paradigm models.} We replace Mamba with the currently popular global ERF RNN paradigm models that have linear computational complexity. The results are shown in Tab.~\ref{tab:rnn}. Under similar FLOPs, \textit{Mamba~\cite{mamba} still demonstrates the best performance and higher throughput.} RWKV6~\cite{rwkv6} achieves results second only to Mamba. In contrast, TTT~\cite{ttt} and xLSTM~\cite{xlstm} fall short in both throughput and performance compared to the former.

\begin{table}[t]
    \centering
    \renewcommand{\arraystretch}{1.0}
    \setlength\tabcolsep{5.0pt}
    \resizebox{1\linewidth}{!}{
        \begin{tabular}{cccccc}
        \toprule[0.1em]
        Model & Reso. & \{$C_1, C_2, C_3$\} & \{$D_1, D_2, D_3$\} & \{$\xi_1, \xi_2, \xi_3$\} & \{$\mu_1, \mu_2, \mu_3$\} \\
    \midrule
    MobileMamba-T2 & 192   & \{144, 272, 368\} & \{1, 2, 2\} & \{0.8, 0.7, 0.6\} & \{0.2, 0.2, 0.3\} \\
    MobileMamba-T4 & 192   & \{176, 368, 448\} & \{1, 2, 2\} & \{0.8, 0.7, 0.6\} & \{0.2, 0.2, 0.3\} \\
    MobileMamba-S6 & 224   & \{192, 384, 448\} & \{1, 2, 2\} & \{0.8, 0.7, 0.6\} & \{0.2, 0.2, 0.3\} \\
    MobileMamba-B1 & 256   & \{200, 376, 448\} & \{2, 3, 2\} & \{0.8, 0.7, 0.6\} & \{0.2, 0.2, 0.3\} \\
    MobileMamba-B2 & 384   & \{200, 376, 448\} & \{2, 3, 2\} & \{0.8, 0.7, 0.6\} & \{0.2, 0.2, 0.3\} \\
    MobileMamba-B4 & 512   & \{200, 376, 448\} & \{2, 3, 2\} & \{0.8, 0.7, 0.6\} & \{0.2, 0.2, 0.3\} \\
        \toprule[0.1em]
        \end{tabular}
    }
    \vspace{-0.2cm}
    \caption{\textbf{Architecture details} of MobileMamba model variants.}
    \label{tab:architecture}
     \vspace{-0.48cm}
\end{table}

\subsection{Training and Testing Strategies}
\label{strategies}
We employ two training strategies to further enhance the performance and efficiency of the small model while maintaining the same number of parameters and computational complexity. Additionally, we use a testing strategy to ensure model effectiveness while improving inference speed. 
\textbf{Knowledge Distillation.} To enable the lightweight student model, MobileMamba, to learn from the more robust teacher classification model, we follow the Soft Distillation setup from DeiT~\cite{training}. This involves minimizing the Kullback-Leibler divergence between the softmax outputs of the teacher model and the student model. 

\noindent\textbf{Extended Training Epochs.} We observe that under the conventional 300 epochs, the loss of the small model MobileMamba has not fully converged, and the Top-1 accuracy has not reached its potential. Therefore, to improve the performance ceiling of the lightweight model, we extend the training to 1000 epochs. 

\noindent\textbf{Normalization Layer Fusion.} Convolution operations are typically followed by batch normalization. During inference, batch normalization can be fused with preceding convolution or linear layers. Recalculating the new convolution layer's weights and biases ensures its combined output matches the original layers' output. This fusion enhances computational efficiency and speeds up the forward pass by reducing the number of layers.

\begin{table}[t]
    \centering
    \renewcommand{\arraystretch}{0.8}
    \setlength\tabcolsep{5.0pt}
    \resizebox{1\linewidth}{!}{
        \begin{tabular}{cccccc}
        \toprule[0.1em]
        Model & FLOPs$\downarrow$ & Params$\downarrow$ & Reso. & Top-1 &\#Pub \\
    \midrule
    EdgeNeXt-XXS~\cite{edgenext} & 260   & 1.3   & 224  & 71.2  & ECCVW'22 \\
    ShuffleNetV2×1.5~\cite{shufflenetv2} & 300   & 3.5   & 224   & 72.6  & ECCV'18 \\
    FasterNet-T0~\cite{fasternet} & 340   & 3.9   & 224  & 71.9  & CVPR'23 \\
    \rowcolor{gray_tab} MobileViTv3-0.5~\cite{mobilevitv3} & 481   & 1.4   & 256   & 72.3  &  arXiv'2209\\
    \rowcolor{gray_tab} EfficientViT-M2~\cite{efficientvit} & 201   & 4.2   & 224   & 70.8  & CVPR'23  \\
    \rowcolor{gray_tab} EMO-1M~\cite{emo} & 261   & 1.3   & 224   & 71.5  & ICCV'23 \\
    \rowcolor{gray_tab} SHViT-S1~\cite{shvit} & 241   & 6.3   & 224   & 72.8  & CVPR'24  \\
    \rowcolor{blue_tab} \textbf{MobileMamba-T2} & 255   & 8.8   & \textbf{192}    & \textbf{73.6} & - \\
    \rowcolor{blue_tab} \textbf{MobileMamba-T2$^\dagger$} & 255   & 8.8   & \textbf{192}  & \textbf{76.9} & - \\
    \midrule
    EdgeNeXt-XS~\cite{edgenext} & 540   & 2.3   & 224  & 75.0  & ECCVW'22 \\
    InceptionNeXt-A~\cite{inceptionnext} & 510   & 4.2   & 224   & 75.3  & CVPR'24 \\
    \rowcolor{gray_tab} EfficientFormerV2-S0~\cite{efficientformerv2} & 400   & 3.5   & 224   & 75.7  & ICCV'23 \\
    \rowcolor{gray_tab} EfficientViT-M4~\cite{efficientvit} & 299   & 8.8   & 224   & 74.3  & CVPR'23  \\
    \rowcolor{gray_tab} EMO-2M~\cite{emo} & 439   & 2.3   & 224   & 75.1  & ICCV'23 \\
    \rowcolor{gray_tab}SHViT-S2~\cite{shvit} & 366   & 11.4  & 224   & 75.2  & CVPR'24  \\
    \rowcolor{oran_tab} VRWKV-T~\cite{vrwkv} & 1200  & 6.2   & 224   & 75.1  & arXiv'2403 \\
    \rowcolor{blue_tab}\textbf{MobileMamba-T4} & 413   & 14.2  & \textbf{192}   & \textbf{76.1} & - \\
    \rowcolor{blue_tab}\textbf{MobileMamba-T4$^\dagger$} & 413   & 14.2  & \textbf{192}   & \textbf{78.9} & - \\
    \midrule
    FasterNet-T1~\cite{fasternet} & 850   & 7.6   & 224  & 76.2  & CVPR'23 \\
    \rowcolor{gray_tab}MobileViTv3-XS~\cite{mobilevitv3} & 927   & 2.5   & 256   & 76.7  & arXiv'2209 \\
    \rowcolor{gray_tab}EfficientViT-M5~\cite{efficientvit} & 522   & 12.4  & 224   & 77.1  & CVPR'23  \\
    \rowcolor{gray_tab}SHViT-S3~\cite{shvit} & 601   & 14.2  & 224   & 77.4  & CVPR'24  \\
    \rowcolor{oran_tab}MSVMamba-N~\cite{msvmamba} & 900   & 7.0   & 224   & 77.3  & NIPS'24 \\
    \rowcolor{oran_tab}Vim-Ti~\cite{vim} & 1500  & 7.0   & 224   & 76.1  & ICML'24 \\
    \rowcolor{oran_tab}LocalVim-T~\cite{localvim} & 1500  & 8.0   & 224   & 76.2  & arXiv'2403 \\
    \rowcolor{oran_tab}EfficientVMamba-T~\cite{efficientvmamba} & 800   & 6.0   & 224   & 76.5  & arXiv'2403 \\
    \rowcolor{blue_tab}\textbf{MobileMamba-S6} & 652   & 15.0  & \textbf{224}   & \textbf{78.0} & - \\
    \rowcolor{blue_tab}\textbf{MobileMamba-S6$^\dagger$} & 652   & 15.0  & \textbf{224}    & \textbf{80.7} & - \\
    \midrule
    MambaOut-Femto~\cite{mambaout} & 1200  & 7.0   & 224  & 78.9  & arXiv'2405 \\
    \rowcolor{gray_tab}MPViT-T~\cite{mpvit} & 1600  & 5.8   & 224   & 78.2  & CVPR'22 \\
    \rowcolor{gray_tab}EMO-6M~\cite{emo} & 961   & 6.1   & 224  & 79.0  & ICCV'23 \\
    \rowcolor{gray_tab}SHViT-S4~\cite{shvit} & 986   & 16.5  & 256   & 79.4  & CVPR'24  \\
    \rowcolor{oran_tab}ViL-T~\cite{vil} & 1500  & 6.0   & 224   & 78.3   & arXiv'2406  \\
    \rowcolor{oran_tab}MSVMamba-M~\cite{msvmamba} & 1500  & 12.0  & 224   & 79.8  & NIPS'24 \\
    \rowcolor{oran_tab}PlainMamba-L1~\cite{plainmamba} & 3000  & 7.0   & 224  & 77.9  & BMVC'24 \\
    \rowcolor{oran_tab}EfficientVMamba-S~\cite{efficientvmamba} & 1300  & 11.0  & 224   & 78.7  & arXiv'2403 \\
    \rowcolor{blue_tab}\textbf{MobileMamba-B1} & 1080  & 17.1  & \textbf{256}  & \textbf{79.9} & - \\
    \rowcolor{blue_tab}\textbf{MobileMamba-B1$^\dagger$} & 1080  & 17.1  & \textbf{256}    & \textbf{82.2} & - \\
    \midrule
    FasterNet-S~\cite{fasternet} & 4560  & 31.1  & 224  & 81.3  & CVPR'23 \\
    \rowcolor{gray_tab}MobileViTv3-0.75~\cite{mobilevitv3} & 2395  & 30.0  & 384   & 77.8  & arXiv'2209 \\
    \rowcolor{gray_tab}MPViT-XS~\cite{mpvit} & 2900  & 10.5  & 224  & 80.9  & CVPR'22 \\
    \rowcolor{gray_tab}EfficientViT-M4r384~\cite{efficientvit} & 1486  & 12.4  & 384   & 79.8   & CVPR'23  \\
    \rowcolor{gray_tab}SHViT-S4r384~\cite{shvit} & 2225  & 16.5  & 384   & 81.0   & CVPR'24   \\
    \rowcolor{oran_tab}ViL-S~\cite{vil} & 5100  & 23.0  & 224   & 81.5   & arXiv'2406  \\
    \rowcolor{oran_tab}VRWKV-S~\cite{vrwkv} & 4600  & 23.8  & 224   & 80.1  & arXiv'2403 \\
    \rowcolor{oran_tab}LocalVim-S~\cite{localvim} & 4800  & 28.0  & 224   & 81.2  & arXiv'2403 \\
    \rowcolor{blue_tab}\textbf{MobileMamba-B2} & 2427  & 17.1  & \textbf{384}   & \textbf{81.6} & - \\
    \rowcolor{blue_tab}\textbf{MobileMamba-B2$^\dagger$} & 2427  & 17.1  & \textbf{384}   & \textbf{83.3} & - \\
    \midrule
    InceptionNeXt-T~\cite{inceptionnext} & 4200  & 28.0  & 224   & 82.3  & CVPR'24 \\
    \rowcolor{gray_tab}MobileViTv3-1.0~\cite{mobilevitv3} & 4220  & 5.1   & 384   & 79.7  & arXiv'2209 \\
    \rowcolor{gray_tab}EfficientViT-M5r512~\cite{efficientvit} & 2670  & 12.4  & 512   & 80.8   &  CVPR'23    \\
    \rowcolor{gray_tab}SHViT-S4r512~\cite{shvit} & 3973  & 16.5  & 512   & 82.0   & CVPR'24   \\
    \rowcolor{oran_tab}ViL-B~\cite{vil} & 18600 & 89.0  & 224   & 82.4   &  arXiv'2406   \\
    \rowcolor{oran_tab}VRWKV-B~\cite{vrwkv} & 18200 & 93.7  & 224   & 82.0  & arXiv'2403 \\
    \rowcolor{oran_tab}PlainMamba-L2~\cite{plainmamba} & 8100  & 25.0  & 224    & 81.6  & BMVC'24 \\
    \rowcolor{oran_tab}Vim-S~\cite{vim} & 5100  & 26.0  & 224   & 80.5  & ICML'24 \\
    \rowcolor{oran_tab}VMamba-T~\cite{vmamba} & 5600  & 22.0  & 224   & 82.2  & NIPS'24 \\
    \rowcolor{oran_tab}EfficientVMamba-B~\cite{efficientvmamba} & 4000  & 33.0  & 224    & 81.8  & arXiv'2403 \\
    \rowcolor{blue_tab}\textbf{MobileMamba-B4} & 4313  & 17.1  & \textbf{512}   & \textbf{82.5} & - \\
    \rowcolor{blue_tab}\textbf{MobileMamba-B4$^\dagger$} & 4313  & 17.1  & \textbf{512}   & \textbf{83.6} & - \\
        \toprule[0.1em]
        \end{tabular}
    }
    \vspace{-0.2cm}
    \caption{\textbf{Classification Performance} on ImageNet-1K~\cite{imagenet} dataset. \protect\sethlcolor{whit_tab}\hl{White}, \protect\sethlcolor{gray_tab}\hl{gray}, \protect\sethlcolor{oran_tab}\hl{yellow}, and \protect\sethlcolor{blue_tab}\hl{blue} backgrounds indicate \textit{CNN-based, Transformer-based, Mamba/RWKV-based} and our MobileMamba, respectively. This kind of display continues for all subsequent experiments. $^\dagger$ indicates the use of training strategies.}
    \label{tab:cls}
    \vspace{-1em}
\end{table}

\section{Experiments} \label{sec:exp}
\vspace{-0.5em}
\subsection{Implementation Details}
We conduct image classification on the ImageNet-1K~\cite{imagenet} dataset. The baseline model is trained from scratch for 300 epochs at a resolution of 224$^2$. The AdamW~\cite{adamw} optimizer is employed with betas (0.9, 0.999), a weight decay of 5e-2, a learning rate of 1.5e-3, and a batch size of 1024. We use a Cosine scheduler~\cite{lrsch} with 20 warmup epochs, Label Smoothing~\cite{labelsmoothing} of 0.1, stochastic depth~\cite{depth}, and RandAugment~\cite{randaugment} during training. For a fair comparison, we follow the same data augmentation techniques proposed in~\cite{training}, including Mixup~\cite{mixup}, random erasing~\cite{erasing}, and auto-augmentation~\cite{autoaugment}.
For the enhanced model$^\dagger$, we train for 1000 epochs and follow the Knowledge Distillation recipe as used in DeiT~\cite{training} with TResNet-L~\cite{tresnet} as the teacher model. 
For object detection tasks, we validate using lightweight SSDLite~\cite{ssdlite} and RetinaNet~\cite{retinanet} on the MS-COCO 2017~\cite{coco} dataset.
For instance segmentation tasks, we conduct experiments using Mask R-CNN~\cite{maskrcnn} on the COCO~\cite{coco} dataset.
For semantic segmentation tasks, we experiment with DeepLabv3~\cite{deeplabv3}, PSPNet~\cite{pspnet}, and FPN~\cite{fpn} on the ADE20K~\cite{ade20k} dataset.
For all downstream task experiments, we use the standard MMDetection~\cite{mmdetection} and MMSegmentation~\cite{mmsegmentation} libraries, and we only replace the optimizer with AdamW~\cite{adamw} without tuning other parameters. GPU throughput is measured on a single Nvidia L40S with a batch size of 256.

\begin{table}[tp]
  \centering
    \resizebox{1\linewidth}{!}{
    \begin{tabular}{cccccccc}
    \toprule[0.17em]
    \multicolumn{8}{c}{Mask R-CNN Object Detection \& Instance Segmentation on COCO} \\
    \midrule
    Backbone & $mAP^b$ & $mAP^b_{50}$ & $mAP^b_{75}$ & $mAP^m$ & $mAP^m_{50}$ & $mAP^m_{75}$ & TP \\
    \midrule
EfficientNet-B0\cite{efficientnet} & 31.9  & 51.0  & 34.5  & 29.4  & 47.9  & 31.2  & 71 \\
 ResNet-50~\cite{resnet} & 38.0  & 58.6  & 41.4  & 34.4  & 55.1  & 36.7  & 41 \\
    \rowcolor{gray_tab}FastViT-SA12~\cite{fastvit} & 38.9  & 60.5  & 42.2  & 35.9  & 57.6  & 38.1  & 36 \\
\rowcolor{gray_tab}EfficientViT-M4\cite{efficientvit} & 32.8  & 54.4  & 34.5  & 31.0  & 51.2  & 32.2  & 121 \\
    \rowcolor{gray_tab}PoolFormer-S12~\cite{metaformer} & 37.3  & 59.0  & 40.1  & 34.6  & 55.8  & 36.9  & 32 \\
    \rowcolor{gray_tab}EfficientFormer-L1~\cite{efficientformer} & 37.9  & 60.3  & 41.0  & 35.4  & 57.3  & 37.3  & 45 \\
    \rowcolor{gray_tab}SHViT-S4~\cite{shvit} & 39.0  & 61.2  & 41.9  & 35.9  & 57.9  & 37.9  & 136 \\
    \rowcolor{gray_tab}EMO-5M~\cite{emo} & 39.3  & 61.7  & 42.4  & 36.4  & 58.4  & 38.7  & 97 \\
    \midrule
    \rowcolor{blue_tab}\textbf{MobileMamba-B1} & \textbf{40.6}  & \textbf{61.8}  & \textbf{43.8}  & \textbf{37.4}  & \textbf{58.9}  & \textbf{39.9}  & \textbf{152} \\
    \toprule[0.12em]
    \multicolumn{8}{c}{RetinaNet Object Detection on COCO} \\
    \toprule[0.17em]
    Backbone & $mAP^b$  & $mAP^b_{50}$ & $mAP^b_{75}$ & $mAP^b_{S}$ & $mAP^b_{M}$ & $mAP^b_{L}$ & TP \\
    \midrule
    MobileNetV3~\cite{mobilenetv3} & 29.9  & 49.3  & 30.8  & 14.9  & 33.3  & 41.1  & 153 \\
    \rowcolor{gray_tab}EfficientViT-M4~\cite{efficientvit} & 32.7  & 52.2  & 34.1  & 17.6  & 35.3  & 46.0  & 160 \\
    \rowcolor{gray_tab}PVTv2-B0~\cite{pvtv2} & 37.2  & 57.2  & 39.5  & 23.1  & 40.4  & 49.7  & 71 \\
    \rowcolor{gray_tab}MobileFormer-508M~\cite{mobileformer} & 38.0  & 58.3  & 40.3  & 22.9  & 41.2  & 49.7  & 58 \\
    \rowcolor{gray_tab}EdgeViT-XXS~\cite{edgevits} & 38.7  & 59.0  & 41.0  & 22.4  & 42.0  & 51.6  & 60 \\
    \rowcolor{gray_tab}SHViT-S4~\cite{shvit} & 38.8  & 59.8  & 41.1  & 22.0  & 42.4  & 52.7  & \textbf{186} \\
    \rowcolor{gray_tab}EMO-5M~\cite{emo} & 38.9  & 59.8  & 41.0    & \textbf{23.8}  & 42.2  & 51.7  & 138 \\
    \rowcolor{oran_tab}EfficientVMamba-T~\cite{efficientvmamba} & 37.5 & 57.8 & 39.6 & 22.6 & 40.7 & 49.1 & 42 \\
    \hline
    \rowcolor{blue_tab}\textbf{MobileMamba-B1} & \textbf{39.6}  & \textbf{59.8}  & \textbf{42.4}  & 21.5  & \textbf{43.1}  & \textbf{53.9}  & 181 \\
    \toprule[0.12em]
    \end{tabular}%
    }
    \vspace{-0.2cm}
    \caption{\textbf{Object Detection} and \textbf{Instance Segmentation} results by RetinaNet~\cite{retinanet} and Mask RCNN~\cite{maskrcnn} on MS-COCO 2017~\cite{coco} dataset. TP: GPU Throughput on a single NVIDIA L40S.}
  \label{tab:objdet}%
  \vspace{-1em}
\end{table}%

\subsection{MobileMamba on ImageNet-1K Classification}
The results of MobileMamba across six different model scales compared to other SoTA methods on ImageNet-1K are presented in Tab.~\ref{tab:cls}. The different model scales are categorized by FLOPs. For instance, compared to the MobileMamba-B1 model, the B2 and B4 models only increase the input resolution without adding network depth or width. In the first two model scales, there are currently no Mamba-based models with equivalent FLOPs. MobileMamba-T2 outperforms Transformer-based SHViT-S1~\cite{shvit} by +0.8$\uparrow$ in Top-1. MobileMamba-T4 surpasses the linear attention-based VRWKV-T~\cite{vrwkv} by +1$\uparrow$ in Top-1 while having only 33\% of its FLOPs. For the MobileMamba-S6 and B1 models, we also observe significant improvements over other CNN, Transformer, and Mamba-based models. MobileMamba-S6 achieves 1.5 higher Top-1 accuracy than EfficientVMamba-T~\cite{efficientvmamba} while reducing FLOPs by 18.5\%$\downarrow$.

To demonstrate the scaling capability of the lightweight models, we maintain the network architecture of the MobileMamba-B1 model and increase the input resolution to 384$^2$ and 512$^2$, resulting in models with about 2G and 4G FLOPs, respectively. The MobileMamba-B2 and B4 models achieve higher classification results while having fewer FLOPs compared to other models. Additionally, the use of training strategies $\dagger$ further enhances model performance. For instance, applying training strategies to the MobileMamba-T2$^\dagger$ model results in a +3.3$\uparrow$ increase in Top-1 and a +1.7$\uparrow$ increase in Top-5. Across all model scales, training strategies consistently demonstrate their ability to significantly improve performance.
\begin{table}[tp]
    \centering
    \renewcommand{\arraystretch}{0.9}
    \setlength\tabcolsep{15.0pt}
    \resizebox{1\linewidth}{!}{
        \begin{tabular}{ccccc}
        \toprule[0.17em]
        Backbone & Reso. & FLOPs $\downarrow$ & \#Params $\downarrow$ & $mAP$ \\
        \hline
        MobileNetv1~\cite{mnetv1}   & 320     & 1.3G  & 5.1        & 22.2 \\
        MobileNetv2~\cite{mnetv2}   & 320        & 0.8G   & 4.3       & 22.1 \\
        MobileNetv3~\cite{mnetv3}   & 320       & 0.6G    & 5.0       & 22.0 \\
        \rowcolor{gray_tab}MobileViTv1-XXS~\cite{mvitv1}  & 320     & 0.9G    & 1.7      & 19.9 \\
        \rowcolor{gray_tab}MobileViTv2-0.5~\cite{mvitv2}  & 320     & 0.9G    & 2.0      & 21.2 \\
        \rowcolor{gray_tab}EMO-1M~\cite{emo}        & 320          & 0.6G     & 2.3     & 22.0 \\
        \rowcolor{blue_tab}\textbf{MobileMamba-B1} & 320 & 1.7G  & 18.0  & \textbf{24.0}  \\
        \hline
        ResNet50~\cite{resnet}     & 512     & 8.8G   & 26.6     & 25.2 \\
        EdgeNeXt-S~\cite{edgenext}  & 512   & 2.1G   & 6.2       & 27.9 \\
        \rowcolor{gray_tab}MobileViTv2-0.75~\cite{mvitv2}   & 512    & 1.8G   & 3.6      & 24.6 \\
        \rowcolor{gray_tab}EMO-2M~\cite{emo}          & 512          & 0.9G   & 3.3     & 25.2\\
        \rowcolor{gray_tab}MobileViTv1-S~\cite{mvitv1}    & 512     & 3.4G    & 5.7      & 27.7 \\
        \rowcolor{gray_tab}MobileViTv2-1.25~\cite{mvitv2}  & 512     & 4.7G   & 8.2      & 27.8 \\
        \rowcolor{gray_tab}EMO-5M~\cite{emo}       & 512        & 1.8G  & 6.0       & 27.9 \\
        \rowcolor{blue_tab}\textbf{MobileMamba-B1} & 512 & 4.4G  & 18.0  & \textbf{29.5}  \\
        \toprule[0.12em]
        \end{tabular}
    }
    \vspace{-0.2cm}
    \caption{\textbf{Object Detection} performance by SSDLite~\cite{mnetv3} on MS-COCO 2017~\cite{coco} dataset at 320$\times$320 resolution.}
    \label{table:det_coco_ssdlite}
    \vspace{-1em}
\end{table}
\subsection{MobileMamba on Downstream Tasks}
\noindent\textbf{Object Detection and Instance Segmentation.} The pre-trained MobileMamba model is evaluated for object detection using light SSDLite~\cite{ssdlite} and heavy RetinaNet~\cite{retinanet}, as well as for instance segmentation with Mask R-CNN~\cite{maskrcnn} on MS-COCO  2017~\cite{coco} dataset. 
For SSDLite in Tab.~\ref{table:det_coco_ssdlite}, we initially experiment at 320$^2$ resolution and subsequently increase to 512$^2$ while keeping other parameters constant. 
MobileMamba-B1 achieves +2$\uparrow$ compared to EMO-1M~\cite{emo} at 320$^2$ resolution. MobileMamba-B1 has -0.3G$\downarrow$ FLOPs than MViTv2-1.25~\cite{mvitv2} while achieving +1.7$\uparrow$ in $mAP$ at 512$^2$. 
For RetinaNet in Tab.~\ref{tab:objdet}, MobileMamba-B1 demonstrates a GPU throughput $\times$4.3$\uparrow$ higher than EfficientVMamba-T~\cite{efficientvmamba}, with +2.1$\uparrow$ in $mAP^b$. Compared to EMO-5M~\cite{emo}, it shows +31\%$\uparrow$ in GPU throughput and +0.7$\uparrow$ in $mAP^b$. 
For Mask R-CNN in Tab.~\ref{tab:objdet}, MobileMamba-B1 shows +57\%$\uparrow$ in throughput compared to EMO-5M~\cite{emo}, with +1.3$\uparrow$ and +1.0$\uparrow$ in $mAP^b$ and $mAP^m$, respectively. Compared to SHViT-S4~\cite{shvit}, it achieves +1.6$\uparrow$ and 1.5$\uparrow$ in $mAP^b$ and $mAP^m$.
\begin{table}[tp]
    \centering
    \renewcommand{\arraystretch}{0.5}
    \setlength\tabcolsep{15.0pt}
    \resizebox{1\linewidth}{!}{
    \begin{tabular}{ccccc}
    \toprule
          & Backbone & FLOPs$\downarrow$ & \#Params.$\downarrow$ & mIoU \\
    \midrule
    \multirow{10}{*}{\rotatebox{90}{\makecell[c]{DeepLabv3\\~\cite{deeplabv3}}}} 
    & MobileNetv2~\cite{mnetv2} & 75.4G & 18.7  & 34.1  \\
    & \cellcolor{gray_tab}MobileNetv2-0.5~\cite{mvitv2} & \cellcolor{gray_tab}26.1G & \cellcolor{gray_tab}6.3   & \cellcolor{gray_tab}31.9  \\
          & \cellcolor{gray_tab}MobileViTv3-0.5~\cite{mvitv3} & \cellcolor{gray_tab}-     & \cellcolor{gray_tab}6.3   & \cellcolor{gray_tab}33.5  \\
          & \cellcolor{gray_tab}EMO-1M~\cite{emo} & \cellcolor{gray_tab}2.4G  & \cellcolor{gray_tab}5.6   & \cellcolor{gray_tab}33.5  \\
          & \cellcolor{gray_tab}MobileViTv2-0.75~\cite{mvitv2} & \cellcolor{gray_tab}40.0G & \cellcolor{gray_tab}9.6   & \cellcolor{gray_tab}34.7  \\
          & \cellcolor{gray_tab}MobileViTv3-0.75~\cite{mvitv3} & \cellcolor{gray_tab}-     & \cellcolor{gray_tab}9.7   & \cellcolor{gray_tab}36.4  \\
          & \cellcolor{gray_tab}EMO-2M~\cite{emo} & \cellcolor{gray_tab}3.5G  & \cellcolor{gray_tab}6.6   & \cellcolor{gray_tab}35.3  \\
\cmidrule{2-5}          & \cellcolor{blue_tab}\textbf{MobileMamba-B4} & \cellcolor{blue_tab}4.7G  & \cellcolor{blue_tab}23.0  & \cellcolor{blue_tab}\textbf{36.6}  \\
    \midrule
    \multirow{16}{*}{\rotatebox{90}{\makecell[c]{Semantic FPN\\~\cite{segfpn}}}} & ResNet-18~\cite{resnet} & 32.2G & 15.5  & 32.9  \\
    & ResNet-50~\cite{resnet} & 45.6G & 28.5  & 36.7  \\
    & ResNet-101~\cite{resnet} & 65.1G & 47.5  & 38.8  \\
          & ResNeXt-101~\cite{resnext} & 64.7G & 47.1  & 39.7  \\
          & \cellcolor{gray_tab}EMO-1M~\cite{emo} & \cellcolor{gray_tab}22.5G & \cellcolor{gray_tab}5.2   & \cellcolor{gray_tab}34.2  \\
          & \cellcolor{gray_tab}PVTv1-Tiny~\cite{pvtv1} & \cellcolor{gray_tab}33.2G & \cellcolor{gray_tab}17.0  & \cellcolor{gray_tab}35.7  \\
          & \cellcolor{gray_tab}PVTv2-BO~\cite{pvtv2} & \cellcolor{gray_tab}25.0G & \cellcolor{gray_tab}7.6   & \cellcolor{gray_tab}37.2  \\
          & \cellcolor{gray_tab}EMO-2M~\cite{emo} & \cellcolor{gray_tab}23.5G & \cellcolor{gray_tab}6.2   & \cellcolor{gray_tab}37.3  \\
          & \cellcolor{gray_tab}PVTv1-Small~\cite{pvtv1} & \cellcolor{gray_tab}44.5G & \cellcolor{gray_tab}28.2  & \cellcolor{gray_tab}39.8  \\
          & \cellcolor{gray_tab}EdgeViT-XXS~\cite{edgevits} & \cellcolor{gray_tab}24.4G & \cellcolor{gray_tab}7.9   & \cellcolor{gray_tab}39.7  \\
          & \cellcolor{gray_tab}EdgeViT-XS~\cite{edgevits} & \cellcolor{gray_tab}27.7G & \cellcolor{gray_tab}10.6  & \cellcolor{gray_tab}41.4  \\
          & \cellcolor{gray_tab}PVTv2-B1~\cite{pvtv2} & \cellcolor{gray_tab}34.2G & \cellcolor{gray_tab}17.8  & \cellcolor{gray_tab}42.5  \\
          & \cellcolor{gray_tab}EMO-5M~\cite{emo} & \cellcolor{gray_tab}25.8G & \cellcolor{gray_tab}8.9   & \cellcolor{gray_tab}40.4  \\
          & \cellcolor{blue_tab}\textbf{MobileMamba-B4} & \cellcolor{blue_tab}\textbf{5.6G}  & \cellcolor{blue_tab}19.8  & \cellcolor{blue_tab}\textbf{42.5}  \\
    \midrule
    \multirow{8}{*}{\rotatebox{90}{\makecell[c]{PSPNet\\~\cite{pspnet}}}} & MobileNetv2~\cite{mnetv2} & 53.1G & 13.7  & 29.7  \\
          & \cellcolor{gray_tab}MobileNetv2-0.5~\cite{mvitv2} & \cellcolor{gray_tab}15.4G & \cellcolor{gray_tab}3.6   & \cellcolor{gray_tab}31.8  \\
          & \cellcolor{gray_tab}EMO-1M~\cite{emo} & \cellcolor{gray_tab}2.1G  & \cellcolor{gray_tab}4.3   & \cellcolor{gray_tab}33.2  \\
          & \cellcolor{gray_tab}MobileViTv2-0.75~\cite{mvitv2} & \cellcolor{gray_tab}26.6G & \cellcolor{gray_tab}6.2   & \cellcolor{gray_tab}35.2  \\
          & \cellcolor{gray_tab}MobileViTv2-1.0~\cite{mvitv2} & \cellcolor{gray_tab}40.3G & \cellcolor{gray_tab}9.4   & \cellcolor{gray_tab}36.5  \\
          & \cellcolor{gray_tab}EMO-2M~\cite{emo} & \cellcolor{gray_tab}3.1G  & \cellcolor{gray_tab}5.5   & \cellcolor{gray_tab}34.5  \\
          & \cellcolor{blue_tab}\textbf{MobileMamba-B4} & \cellcolor{blue_tab}4.5G  & \cellcolor{blue_tab}20.5  & \cellcolor{blue_tab}\textbf{36.9}  \\
    \bottomrule 
    \end{tabular}%
    }
    \vspace{-0.2cm}
    \caption{\textbf{Semantic Segmentation} results by DeepLabv3~\cite{deeplabv3}, Semantic FPN~\cite{segfpn}, and PSPNet~\cite{pspnet} on ADE20K~\cite{ade20k} dataset at 512$\times$512 resolution.}
    \label{table:seg_three}
    \vspace{-1em}
\end{table}%

\noindent\textbf{Semantic Segmentation.} The pre-trained MobileMamba is evaluated for semantic segmentation performance using DeepLabv3~\cite{deeplabv3}, Semantic FPN~\cite{segfpn}, and PSPNet~\cite{pspnet} on the ADE20K~\cite{ade20k} dataset in Tab.~\ref{table:seg_three}.
For DeepLabv3, MobileMamba-B4 achieves +1.3$\uparrow$ in mIoU over EMO-2M~\cite{emo}.
For Semantic FPN, MobileMamba demonstrates a significant advantage. Compared to EMO-5M~\cite{emo}, it has only 22\% of the FLOPs while +2.1$\uparrow$ in mIoU.
Compared to EdgeViT-XS~\cite{edgevits}, it achieves +1.1$\uparrow$ mIoU with only 20\% of the FLOPs. Compared to PVTv2-B1~\cite{pvtv2} with similar results, the FLOPs of our model are only 16\% of theirs.
For PSPNet, MobileMamba-B4 achieves +2.4$\uparrow$ in mIoU over EMO-2M~\cite{emo}.
Compared to MobileViTv2-1.0~\cite{mvitv2}, it achieves +0.4$\uparrow$ higher mIoU with only 11\% of the FLOPs.
\subsection{Extra Ablation and Explanatory Analysis}
\label{ablation}
\begin{figure}[t]
\vspace{-1em}
\centering
\includegraphics[width=0.47\textwidth]{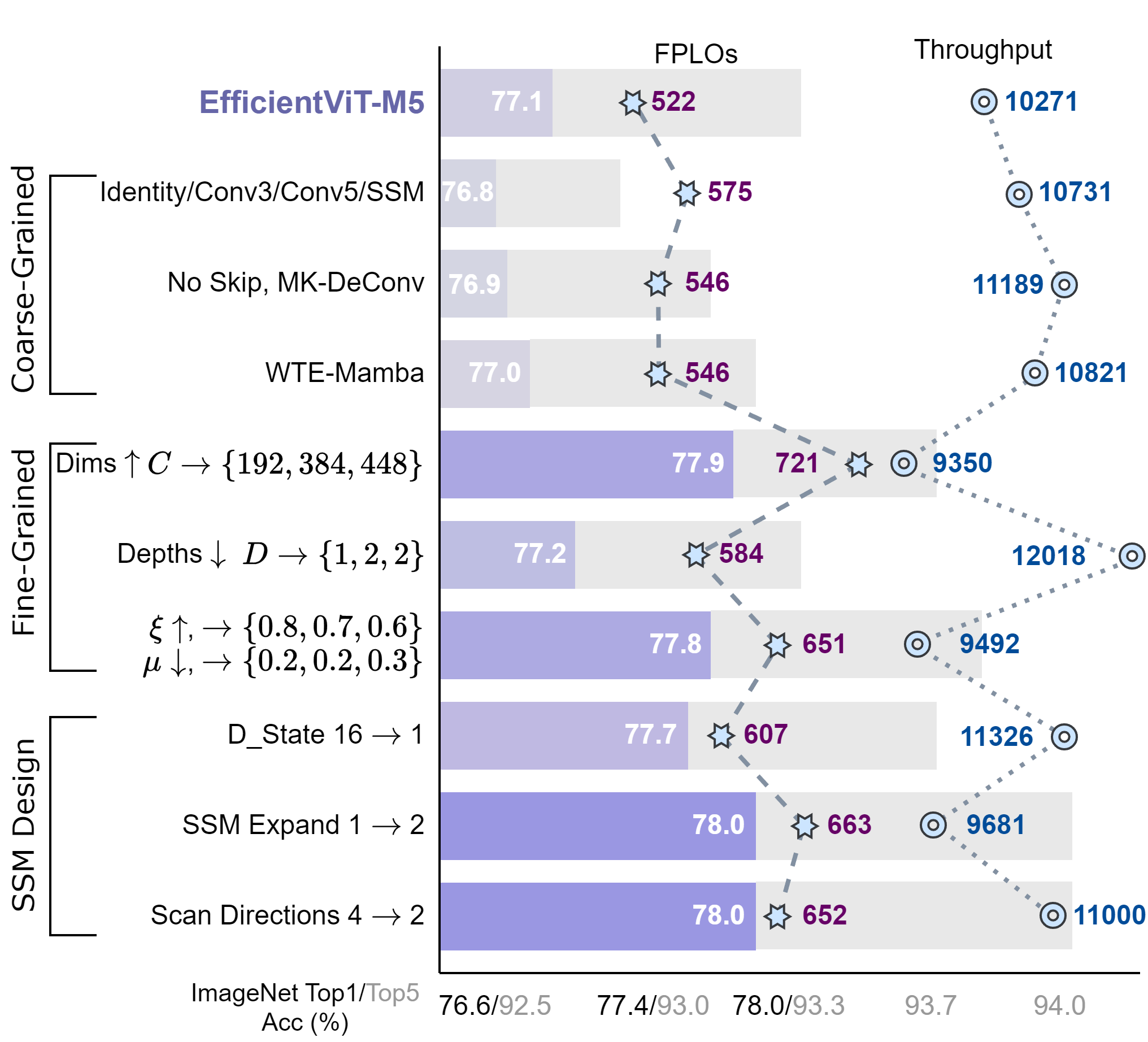}
\caption{\textbf{Incremental Experiments} on the ImageNet-1K for MobileMamba compare Top1/Top5 Acc., FLOPs, and Throughput.}
\label{fig:increamental}
\vspace{-1em}
\end{figure}

\noindent\textbf{Incremental Experiments.} Fig.~\ref{fig:increamental} illustrates the process of deriving the MobileMamba model from the baseline EfficientViT-M5~\cite{efficientvit} model through incremental experiments. 
Since FLOPs do not always fully reflect the inference speed of a model, we include the GPU throughput metric to demonstrate the model's efficiency.
At the structural level, the Cascaded Group Attention is gradually transformed into the proposed MRFFI module. This process integrates multi-scale receptive fields while enhancing the model's throughput. Subsequently, a Fine-Grained design is applied to enhance the model's representation capabilities in terms of Mamba's global receptive field and frequency domain details, thereby improving model performance. \textit{Reducing the number of network layers while increasing the dimension and global proportion $\xi$ can decrease FLOPs and simultaneously increase throughput with similar accuracy.} Finally, \textit{by decreasing d\_state and scan directions and increasing the expanding ratio of SSM, the model's performance is further enhanced, and its throughput is significantly increased.}
Ultimately, compared to the baseline, MobileMamba achieves an improvement of +0.9$\uparrow$ in Top-1 and +0.6$\uparrow$ in Top-5 accuracy, while also increasing throughput by +729$\uparrow$ images per second.
\begin{table}[bp]
\vspace{-0.2cm}
  \centering
  \renewcommand{\arraystretch}{1.0}
    \setlength\tabcolsep{5.0pt}
    \resizebox{1\linewidth}{!}{
    \begin{tabular}{ccccccccc}
    \toprule
    \multirow{2}[4]{*}{Model} & \multicolumn{1}{c}{FLOPs} & \multicolumn{1}{c}{\#Params} & \multirow{2}[4]{*}{Reso.} & \multicolumn{2}{c}{Throughput} & \multicolumn{2}{c}{Latency} & \multicolumn{1}{c}{\multirow{2}[4]{*}{Top-1\newline{}(\%)}} \\
\cmidrule{5-8}          &   (M)    &   (M)    &       & GPU   & CPU   & CPU   & Mobile &  \\
    \midrule
    \rowcolor{gray_tab}EfficientViT-M2~\cite{efficientvit} & 201   & 4.2   & 224   & 18693  & 255   & 3.9   & 1.1   & 70.8  \\
    \rowcolor{gray_tab}EMO-1M~\cite{emo} & 261   & 1.3   & 224   & 8361  & 91    & 10.9  & 5.1   & 71.5  \\
    \rowcolor{blue_tab}\textbf{MobileMamba-T2} & 255   & 8.8   & 192   & 21071  & 85    & 11.8  & 11.7  & 73.6  \\
    \midrule
    \rowcolor{gray_tab}EfficientViT-M4~\cite{efficientvit} & 299   & 8.8   & 224   & 14612  & 228   & 4.4   & 1.6   & 74.3  \\
    \rowcolor{gray_tab}EMO-2M~\cite{emo} & 439   & 2.3   & 224   & 6301  & 67    & 15.0  & 7.6   & 75.1  \\
    \rowcolor{blue_tab}\textbf{MobileMamba-T4} & 413   & 14.2  & 192   & 16571  & 84    & 11.8  & 16.9  & 76.1  \\
    \midrule
    \rowcolor{gray_tab}EfficientViT-M5~\cite{efficientvit} & 522   & 12.4  & 224   & 10271  & 180   & 5.6   & 2.0   & 77.1  \\
    \rowcolor{oran_tab}EfficientVMamba-T~\cite{efficientvmamba} & 800   & 6.0   & 224   & 6285  & 14    & 70.0  & 113.6  & 76.5  \\
    \rowcolor{blue_tab}\textbf{MobileMamba-S6} & 652   & 15.0  & 224   & 11000  & 80    & 12.5  & 19.7  & 78.0  \\
    \midrule
    \rowcolor{oran_tab}EfficientVMamba-S~\cite{efficientvmamba} & 1300  & 11.0  & 224   & 3327  & 7     & 137.8  & 287.9  & 78.7  \\
    \rowcolor{gray_tab}EMO-6M~\cite{emo} & 961   & 6.1   & 224   & 4038  & 42    & 23.5  & 14.8  & 79.0  \\
    \rowcolor{blue_tab}\textbf{MobileMamba-B1} & 1080  & 17.1  & 256   & 6986  & 49    & 20.3  & 47.0  & 79.9  \\
    \midrule
    \rowcolor{gray_tab}EfficientViT-M5r512~\cite{efficientvit} & 2670  & 12.4  & 512   & 1694  & 31    & 32.4  & 3.5   & 80.8  \\
    \rowcolor{oran_tab}EfficientVMamba-B~\cite{efficientvmamba} & 4000  & 33.0  & 224   & 648   & 5     & 198.2  & 834.8  & 81.8  \\
    \rowcolor{blue_tab}\textbf{MobileMamba-B4} & 4313  & 17.1  & 512   & 1862  & 12    & 84.2  & 291.7  & 82.5  \\
    \bottomrule
    \end{tabular}%
    }
    \caption{Comparison with SoTAs in \textbf{Effectiveness} and \textbf{Efficiency}.}
  \label{tab:efficiency}%
\end{table}%

\noindent\begin{table*}[tp]
    \centering
    \subfloat[Small model for Lower Resolution.]{
        \label{tab:lowreso}
        \renewcommand{\arraystretch}{0.9}
        \setlength\tabcolsep{6.0pt}
        \resizebox{0.3\linewidth}{!}{
        \begin{tabular}{cccccc}
        \toprule[0.1em]
        Reso. & FLOPs & Throughput & Params & Top-1 \\
    \midrule
    160   & 252   & \textbf{21548} & 11.7  & 72.6  \\
    192   & 255   & 21071  & 8.8   & \textbf{73.5} \\
    224   & 269   & 20203  & 6.5   & 73.3  \\
    \bottomrule[0.10em]
        \end{tabular}
    }
    } \hfill
    \subfloat[Large model for Higher Resolution.]{
        \label{tab:highreso}
        \renewcommand{\arraystretch}{0.8}
        \setlength\tabcolsep{6.5pt}
        \resizebox{0.27\linewidth}{!}{
        \begin{tabular}{cccccc}
        \toprule[0.1em]
        Reso. & FLOPs & Throughput & Params & Top-1 \\
    \midrule
    224   & 2305  & \textbf{3602}  & 50.7  & 80.0  \\
    384   & 2427  & 3175  & 17.1  & \textbf{81.6} \\
    \midrule
    224   & 4374  & \textbf{2145}  & 79.4  & 80.7  \\
    512   & 4313  & 1862  & 17.1  & \textbf{82.5} \\
        \toprule[0.1em]
        \end{tabular}
    }
    } \hfill
    \subfloat[Training and Testing Strategies.]{
        \label{tab:strategy}
        \renewcommand{\arraystretch}{0.5}
        \setlength\tabcolsep{7.0pt}
        \resizebox{0.28\linewidth}{!}{
        \begin{tabular}{ccccc}
        \toprule[0.1em]
        Method & FLOPs & Throughput & Top-1 \\
    \midrule
    S6 & 652   & 9200  & 78.0  \\
    +KD   & 652   & 9200  & 80.0  \\
    +1000e & 652   & 9200  & 80.7  \\
    +NLF & 648   & \textbf{11000} & \textbf{80.7} \\
        \toprule[0.1em]
        \end{tabular}
    }
    } \hfill
    \vspace{0.5em}
    \subfloat[Impact of Mamba Component.]{
        \label{tab:mamba}
        \renewcommand{\arraystretch}{0.6}
        \setlength\tabcolsep{12.0pt}
        \resizebox{0.4\linewidth}{!}{
        \begin{tabular}{cccccccc}
        \toprule[0.1em]
        S & R & D & FLOPs & Throughput & Params & Top-1 \\
    \midrule
    4     & 1     & 1     & 607   & \textbf{11782} & 14.1  & 77.7  \\
    4     & 1     & 16    & 651   & 9945  & 14.3  & 77.8 \\
    4     & 2     & 1     & 663   & 10069  & 15.2  & \textbf{78.1} \\
    2     & 1     & 16    & 624   & 11121  & 14.1  & 77.6 \\
    2     & 2     & 1     & 652   & 11000  & 15.0  & 78.0 \\
        \toprule[0.1em]
        \end{tabular}
    }
    } \hfill
    \subfloat[Impact of MK-DeConv.]{
        \label{tab:cnn}
        \renewcommand{\arraystretch}{0.5}
        \setlength\tabcolsep{10.0pt}
        \resizebox{0.20\linewidth}{!}{
        \begin{tabular}{cccc}
    \toprule[0.1em]
    $n$. & TP & Top-1 \\
    \midrule
    1   & 11000  & 78.0  \\
    2   & 10847  & 77.9  \\
    3   & 10791  & 78.0  \\
        \toprule[0.1em]
        \end{tabular}
    }
    } \hfill
    \subfloat[Impact of Wave Transformation.]{
        \label{tab:wt}
        \renewcommand{\arraystretch}{1.3}
        \setlength\tabcolsep{7.0pt}
        \resizebox{0.3\linewidth}{!}{
        \begin{tabular}{ccccc}
    \toprule[0.1em]
    Method & Params & FLOPs & Throughput & Top-1 \\
    \midrule
    wo WT & 14.9  & 652   & \textbf{11687} & 77.8 \\
    w WT  & 15.0  & 652   & 11000  & \textbf{78.0}  \\
        \toprule[0.1em]
        \end{tabular}
    }
    } \hfill
    \vspace{-0.2cm}
    \caption{\textbf{Ablation Studies} and comparison analysis on ImageNet-1K~\cite{imagenet}.}
    \label{table:ablation_all}
    \vspace{-1.0em}
\end{table*}
\noindent\textbf{Efficiency Comparison.} Tab.~\ref{tab:efficiency} presents a comparison with SoTA methods in efficiency and effectiveness. MobileMamba surpasses all methods in GPU throughput. On average, the three different sizes of MobileMamba models achieve $\times3.5\uparrow$ faster in GPU throughput compared to EfficientVMamba~\cite{efficientvmamba}.
However, for CPU throughput on AMD EPYC 9K84 96-Core and latency and latency on the mobile iPhone15 (ms), MobileMamba lags behind Transformer-based models. \textit{This is attributed to the current engineering implementation of the Mamba method on CPUs, which still has room for improvement and optimization.} Nevertheless, compared to other Mamba-based methods, MobileMamba achieves only 15\%-42\% latency over EfficientVMamba on CPUs, while also achieving an average improvement of +1.5$\uparrow$ in Top-1.

\noindent\textbf{Ablations on Small Models with Low-Resolution.}
To enhance the performance of smaller models while increasing their throughput, we investigate the impact of input resolution. We set three input resolutions: $160^2$, $192^2$, and $224^2$, and adjust the model parameters to ensure that the FLOPs are approximately \textit{250M} for each resolution. As shown in Tab.~\ref{tab:lowreso}, \textit{despite similar FLOPs, lower input resolutions result in higher model throughput and larger parameter sizes}. Considering throughput, parameter size, and performance, we design the small model with an input resolution of $192^2$, achieving a good balance and satisfactory results.

\noindent\textbf{Ablations on Large Models with High-Resolution.}
We explore ways to enhance the scaling capability of small models. In Tab.~\ref{tab:highreso}, at the standard resolution of $224^2$, increasing the model's depth and width to achieve \textit{2G} and \textit{4G} FLOPs does not significantly improve performance despite the increased computational load. \textit{This is due to the excessively low input resolution in the current three-stage framework}. Therefore, we increase the input resolutions to $384^2$ and $512^2$. With similar FLOPs and a slight loss in throughput, the Top-1 improved by +1.6$\uparrow$ and +1.8$\uparrow$, respectively.

\noindent\textbf{Effect of Training Strategies.}
Tab.~\ref{tab:strategy} presents the incremental experiments using training and testing strategies. After applying KD, there is an increase of +2$\uparrow$ in Top-1 and +0.7$\uparrow$ in Top-5 accuracy on the ImageNet-1K dataset. Extending the training to 1000 epochs further improves these metrics by +0.7$\uparrow$ and +0.5$\uparrow$, respectively. Ultimately, the model with \textit{652M} FLOPs achieves results of 80.7 in Top-1 and 95.2 in Top-5 accuracy, surpassing the model with \textit{1080M} FLOPs that did not use the training strategies. Additionally, employing normalization fusion during the testing phase can further enhance the speed by $\times1.2\%\uparrow$.

\noindent\textbf{Ablations on Mamba Component.}
Experiments on the internal parameters of the Mamba model are shown in Tab.~\ref{tab:mamba}. S, R, and D represent \textit{scanning directions}, \textit{expanding ratios} and \textit{d\_state}, respectively. Reducing the S can increase throughput, albeit with a slight decrease in performance. With the same number of S, using R=2 and D=1 results in higher throughput and better performance compared to R=1 and D=16. Therefore, the final choice is to use bidirectional scanning with an R=2 and D=1.

\noindent\textbf{Ablations on MK-DeConv.}
We experimented with the number of splits \( n \) in the efficient MK-DeConv operation (see Tab.~\ref{tab:cnn}). For \( n=1 \), all channels use a single convolution module with a kernel size of 3. For \( n=3 \), channels are split into three groups with $k=3,5,7$ respectively and then concatenated along the channel dimension. The methods show no significant differences in parameters, FLOPs, and throughput yielding similar results. Thus, we adopt \( n=1 \) for simplicity.
However, using $k=3$ results in an \textit{ERF} of 3. After WT, the feature map size is halved, followed by convolution with the same $k=3$ and an IWT, restoring the original feature size and effectively doubling the \textit{ERF} to 6. \textit{This approach achieves multi-kernel and multi-receptive field characteristics by combining single-branch convolutions and wavelet transformations.}

\noindent\textbf{Effect of Wave Transformation Component.} The wavelet transform generates one low-frequency and three high-frequency feature maps. The low-frequency map retains the original feature information, while the high-frequency maps capture edge details. Post wavelet transform, the halved feature maps undergo convolution and then inverse wavelet transform, restoring the original size and effectively doubling the receptive field. Despite a potential reduction in throughput, \textit{the wavelet transform's benefits in enhancing the receptive field and extracting edge information can improve model performance} (see Tab.~\ref{tab:wt}).

\vspace{-0.5em}
\section{Conclusion} \label{sec:conclusion}
We designed the MobileMamba framework to balance performance and efficiency, addressing the limitations of existing Mamba-based models. The proposed MRFFI module enhances perception across various receptive fields while preserving high-frequency features and inference efficiency. The training and testing strategies further enhance performance and efficiency. Extensive experiments on ImageNet-1K dataset validate the method's effectiveness, efficiency, and transferability in high-resolution downstream tasks.

\noindent\textbf{Limitations and Future Work.} 
Mamba models, despite their advancements, still exhibit engineering implementation shortcomings, including the need for substantial improvements in CPU acceleration and edge device acceleration. In the future, we will continue to concentrate on enhancing the inference capabilities of Mamba models across a range of devices, with a particular focus on efficiency.
{
    \small
    \bibliographystyle{ieeenat_fullname}
    \bibliography{main}
}
\clearpage
\renewcommand\thefigure{A\arabic{figure}}
\renewcommand\thetable{A\arabic{table}}  
\renewcommand\theequation{A\arabic{equation}}
\setcounter{equation}{0}
\setcounter{table}{0}
\setcounter{figure}{0}
\appendix

\section*{Supplementary Material Overview}
The supplementary material presents more comprehensive analysis and results of our MobileMamba to facilitate the comparison of subsequent methods:
\begin{itemize}
    \item \textbf{Sec.~\ref{sec:supp_finegrain}} provides more detailed Fine-Grained design analysis and experiments on ImageNet-1K~\cite{imagenet} dataset.
    \item \textbf{Sec.~\ref{sec:supp_kernel}} provides more detailed Kernel Size analysis and experiments on ImageNet-1K~\cite{imagenet} dataset.
    \item \textbf{Sec.~\ref{sec:supp_droppath}} provides more detailed DropPath analysis and experiments on ImageNet-1K~\cite{imagenet} dataset.
    \item \textbf{Sec.~\ref{sec:supp_viserf}} provides more detailed ERF Visualization analysis compared with different structure SoTA methods on ImageNet-1K~\cite{imagenet} dataset.
    \item \textbf{Sec.~\ref{sec:supp_analydown}} provides more detailed Pre-trained Models with Different Resolutions for Downstream Tasks analysis and experiments on MS-COCO 2017~\cite{coco} and ADE20K~\cite{ade20k} dataset.
    \item \textbf{Sec.~\ref{sec:supp_det}} provides more detailed object detection results using different frameworks on MS-COCO 2017~\cite{coco} dataset. 
    \item \textbf{Sec.~\ref{sec:supp_seg}} provides more detailed semantic segmentation results using Mask R-CNN~\cite{maskrcnn} for multiple magnitudes of MobileMamba on ADE20K~\cite{ade20k} dataset. 
    \item The \textbf{Codes} folder in the supplementary materials contains all the training and testing code for the models, as well as the log files for each model.
\end{itemize}
\section{More Ablation and Explanatory Analysis}
\subsection{Fine-Grained design analysis} 
\label{sec:supp_finegrain}
We conducted experiments to analyze the impact of global and local channel ratios in Tab.~\ref{tab:sup_gl}, dimensions in Tab.~\ref{tab:sup_dim}, and depth in Tab.~\ref{tab:sup_depth}. For the global and local channel ratios, we observed the importance of global channels for model performance, despite a slight decrease in throughput. In higher stages, due to the increased number of channels, some redundancy may exist. To reduce computational load, we directly map 10\% of the channels in the last two stages.

Regarding dimensionality, we controlled variables by maintaining similar FLOPs and throughput while adjusting the global and local ratios to accommodate different dimensional changes. Altering the dimensions in stage 1 significantly affects FLOPs and throughput, whereas changes in stage 3 primarily impact the number of model parameters. To maximize dimensions in each stage while maintaining low FLOPs and high throughput, we selected \{192, 384, 448\} as the dimensions for each stage.

For model depth, we found that increasing depth significantly reduces throughput. Therefore, we increased depth while maintaining similar throughput, but the effect was limited due to lower FLOPs under the same conditions. In extreme cases, where each stage has only one layer and FLOPs are balanced with other models, throughput is significantly higher, but performance is poor. After trade-offs, we chose a depth of \{1, 2, 2\}.

\begin{table}[htp]
\centering
\caption{Ablations on Global $\xi$ and Local $\mu$ Ratios.}
\label{tab:sup_gl}
\resizebox{1\linewidth}{!}{
\begin{tabular}{cccccc}
\toprule
\begin{tabular}[c]{@{}c@{}}\(\{C_1, C_2, C_3\}\)
\\      \(\{D_1, D_2, D_3\}\)\end{tabular}                       & \begin{tabular}[c]{@{}c@{}}\(\{\xi_1, \xi_2, \xi_3\}\) \\      \(\{\mu_1, \mu_2, \mu_3\}\)\end{tabular}                              & \begin{tabular}[c]{@{}c@{}}FLOPs\\      (M)\end{tabular} & \begin{tabular}[c]{@{}c@{}}Params\\      (M)\end{tabular} & Throughput & Top-1 \\
\midrule
\begin{tabular}[c]{@{}c@{}}\{192,   384, 448\}\\      \{1, 2, 2\}\end{tabular} & \begin{tabular}[c]{@{}c@{}}\{0.6, 0.6, 0.6\}\\      \{0.4, 0.3, 0.3\}\end{tabular} & 620                                                      & 14.6                                                      & 11353      & 77.5  \\
\begin{tabular}[c]{@{}c@{}}\{192,   384, 448\}\\      \{1, 2, 2\}\end{tabular} & \begin{tabular}[c]{@{}c@{}}\{0.7, 0.6, 0.5\}\\      \{0.2, 0.2, 0.3\}\end{tabular} & 619                                                      & 14.3                                                      & 11815      & 77.7  \\
\begin{tabular}[c]{@{}c@{}}\{192, 384, 448\}\\      \{1, 2, 2\}\end{tabular}   & \begin{tabular}[c]{@{}c@{}}\{0.8, 0.6, 0.6\}\\      \{0.2, 0.3, 0.3\}\end{tabular} & 637                                                      & 14.7                                                      & 11222      & 77.7  \\
\begin{tabular}[c]{@{}c@{}}\{192,   384, 448\}\\      \{1, 2, 2\}\end{tabular} & \begin{tabular}[c]{@{}c@{}}\{0.8, 0.7, 0.6\}\\      \{0.2, 0.3, 0.4\}\end{tabular} & 652                                                      & 15.0                                                      & 10949      & 77.8  \\
\begin{tabular}[c]{@{}c@{}}\{192,   384, 448\}\\      \{1, 2, 2\}\end{tabular} & \begin{tabular}[c]{@{}c@{}}\{0.8, 0.8, 0.8\}\\      \{0.2, 0.1, 0.1\}\end{tabular} & 675                                                      & 16.0                                                      & 10560      & 78.0  \\
\begin{tabular}[c]{@{}c@{}}\{192,   384, 448\}\\      \{1, 2, 2\}\end{tabular} & \begin{tabular}[c]{@{}c@{}}\{0.0, 0.7, 0.8\}\\      \{1.0, 0.2, 0.1\}\end{tabular} & 618                                                      & 14.8                                                      & 12735      & 77.2  \\
\begin{tabular}[c]{@{}c@{}}\{192,   384, 448\}\\      \{1, 2, 2\}\end{tabular} & \begin{tabular}[c]{@{}c@{}}\{0.6, 0.7, 0.8\}\\      \{0.4, 0.2, 0.1\}\end{tabular} & 646                                                      & 15.6                                                      & 11546      & 77.8  \\
\begin{tabular}[c]{@{}c@{}}\{192,   384, 448\}\\      \{1,2,2\}\end{tabular}   & \begin{tabular}[c]{@{}c@{}}\{0.8, 0.7, 0.6\}\\      \{0.2, 0.2, 0.3\}\end{tabular} & 652                                                      & 15.0                                                      & 11000      & 78.0 \\
\bottomrule
\end{tabular}}
\end{table}

\begin{table}[htp]
\centering
\caption{Ablations on Dimensions $C$.}
\label{tab:sup_dim}
\resizebox{1\linewidth}{!}{
\begin{tabular}{cccccc}
\toprule
\begin{tabular}[c]{@{}c@{}}\(\{C_1, C_2, C_3\}\)
\\      \(\{D_1, D_2, D_3\}\)\end{tabular}                       & \begin{tabular}[c]{@{}c@{}}\(\{\xi_1, \xi_2, \xi_3\}\) \\      \(\{\mu_1, \mu_2, \mu_3\}\)\end{tabular}                              & \begin{tabular}[c]{@{}c@{}}FLOPs\\      (M)\end{tabular} & \begin{tabular}[c]{@{}c@{}}Params\\      (M)\end{tabular} & Throughput & Top-1 \\
\midrule
\begin{tabular}[c]{@{}c@{}}\{192,   320, 368\}\\      \{1, 3, 4\}\end{tabular} & \begin{tabular}[c]{@{}c@{}}\{0., 0.75, 0.75\}\\      \{0.9, 0.15, 0.15\}\end{tabular}    & 632                                                      & 15.9                                                      & 10811      & 77.7  \\
\begin{tabular}[c]{@{}c@{}}\{172,   320, 368\}\\      \{1, 3, 4\}\end{tabular} & \begin{tabular}[c]{@{}c@{}}\{0.65, 0.65, 0.65\}\\      \{0.25, 0.25, 0.25\}\end{tabular} & 581                                                      & 15.1                                                      & 10809      & 77.6  \\
\begin{tabular}[c]{@{}c@{}}\{180,   336, 368\}\\      \{1, 3, 4\}\end{tabular} & \begin{tabular}[c]{@{}c@{}}\{0.5, 0.5, 0.5\}\\      \{0.4, 0.4, 0.4\}\end{tabular}       & 595                                                      & 14.8                                                      & 10916      & 77.6  \\
\begin{tabular}[c]{@{}c@{}}\{192,   336, 368\}\\      \{1, 3, 4\}\end{tabular} & \begin{tabular}[c]{@{}c@{}}\{0.4, 0.4, 0.4\}\\      \{0.5, 0.5, 0.5\}\end{tabular}       & 599                                                      & 14.2                                                      & 11373      & 77.5  \\
\begin{tabular}[c]{@{}c@{}}\{208, 400, 464\}\\      \{1, 2, 2\}\end{tabular}   & \begin{tabular}[c]{@{}c@{}}\{0.6, 0.5, 0.4\}\\      \{0.3, 0.4, 0.5\}\end{tabular}       & 678                                                      & 15.2                                                      & 11120      & 77.8  \\
\begin{tabular}[c]{@{}c@{}}\{224,   336, 400\}\\      \{1,2,2\}\end{tabular}   & \begin{tabular}[c]{@{}c@{}}\{0.8,0.7,0.6\}\\      \{0.2,0.2,0.3\}\end{tabular}           & 659                                                      & 12.6                                                      & 11337      & 77.4  \\
\begin{tabular}[c]{@{}c@{}}\{208,   384, 416\}\\      \{1,2,2\}\end{tabular}   & \begin{tabular}[c]{@{}c@{}}\{0.8,0.7,0.6\}\\      \{0.2,0.2,0.3\}\end{tabular}           & 681                                                      & 14.5                                                      & 10956      & 78.1  \\
\begin{tabular}[c]{@{}c@{}}\{176,   384, 480\}\\      \{1,2,2\}\end{tabular}   & \begin{tabular}[c]{@{}c@{}}\{0.8,0.7,0.6\}\\      \{0.2,0.2,0.3\}\end{tabular}           & 622                                                      & 15.7                                                      & 11599      & 77.8  \\
\begin{tabular}[c]{@{}c@{}}\{192,   384, 448\}\\      \{1,2,2\}\end{tabular}   & \begin{tabular}[c]{@{}c@{}}\{0.8,0.7,0.6\}\\      \{0.2,0.2,0.3\}\end{tabular}           & 652                                                      & 15.0                                                      & 11000      & 78.0 \\
\bottomrule
\end{tabular}}
\end{table}

\begin{table}[htp]
\centering
\caption{Ablations on Depth $D$.}
\label{tab:sup_depth}
\resizebox{1\linewidth}{!}{
\begin{tabular}{cccccc}
\toprule
\begin{tabular}[c]{@{}c@{}}\(\{C_1, C_2, C_3\}\)
\\      \(\{D_1, D_2, D_3\}\)\end{tabular}                       & \begin{tabular}[c]{@{}c@{}}\(\{\xi_1, \xi_2, \xi_3\}\) \\      \(\{\mu_1, \mu_2, \mu_3\}\)\end{tabular}                              & \begin{tabular}[c]{@{}c@{}}FLOPs\\      (M)\end{tabular} & \begin{tabular}[c]{@{}c@{}}Params\\      (M)\end{tabular} & Throughput & Top-1 \\
\midrule
\begin{tabular}[c]{@{}c@{}}\{160,   304, 448\}\\      \{2, 3, 3\}\end{tabular} & \begin{tabular}[c]{@{}c@{}}\{0.15, 0.55, 0.55\}\\      \{0.35, 0.35, 0.35\}\end{tabular} & 560                                                      & 14.8                                                      & 11236      & 77.6  \\
\begin{tabular}[c]{@{}c@{}}\{128,   256, 384\}\\      \{3, 4, 5\}\end{tabular} & \begin{tabular}[c]{@{}c@{}}\{0.15, 0.55, 0.55\}\\      \{0.35, 0.35, 0.35\}\end{tabular} & 510                                                      & 15.0                                                      & 11155      & 77.1  \\
\begin{tabular}[c]{@{}c@{}}\{192,   384, 448\}\\      \{1,2,2\}\end{tabular}   & \begin{tabular}[c]{@{}c@{}}\{0.8,0.7,0.6\}\\      \{0.2,0.2,0.3\}\end{tabular}           & 652                                                      & 15.0                                                      & 11006      & 78.0  \\
\begin{tabular}[c]{@{}c@{}}\{208,   416, 624\}\\      \{1, 1, 1\}\end{tabular} & \begin{tabular}[c]{@{}c@{}}\{0.8, 0.7, 0.6\}\\      \{0.2, 0.2, 0.3\}\end{tabular}       & 648                                                      & 15.3                                                      & 12397      & 77.4  \\
\begin{tabular}[c]{@{}c@{}}\{192,   384, 576\}\\      \{1, 2, 1\}\end{tabular} & \begin{tabular}[c]{@{}c@{}}\{0.8, 0.7, 0.6\}\\      \{0.2, 0.2, 0.3\}\end{tabular}       & 651                                                      & 15.1                                                      & 11000      & 77.8  \\
\bottomrule
\end{tabular}}
\end{table}

\subsection{Effect of kernel sizes} 
\label{sec:supp_kernel}
We experimented with the impact of different convolution kernel sizes across stages, as shown in Tab.~\ref{tab:sup_kernel}. Using the same kernel size across different stages yields similar results. However, reducing the kernel size as the feature map scale decreases with increasing stages improves model performance.

\begin{table}[htp]
    \centering
    \caption{Ablations on Kernel Size}
    \label{tab:sup_kernel}
    \renewcommand{\arraystretch}{0.8}
    \setlength\tabcolsep{6.0pt}
    \resizebox{1.0\linewidth}{!}{
\begin{tabular}{ccccc}
    \toprule[0.17em]
    Size   & FLOPs(M) & Params(M) & Throughput & Top-1         \\
    \hline
    \{7,7,7\} & 15.2     & 653       & 10937      & 77.7          \\
    \{5,5,3\} & 15.0       & 652       & 11142      & 77.8          \\
    \{5,3,3\} & 15.0       & 652       & 11130      & 77.6          \\
    \{7,5,3\} & 15.0     & 652       & 11000      &78.0 \\
        \bottomrule[0.12em]
        \end{tabular}
    }
\end{table}

\subsection{Effect of DropPath} 
\label{sec:supp_droppath}
For the MobileMamba-T2, T4, and S6 models, we did not use DropPath due to their shallow depth. In the B1 model, we applied DropPath, with specific results shown in Tab.~\ref{tab:sup_dp}. A DropPath value of 0.03 achieved the best performance, increasing Top-1 accuracy by 0.2 compared to not using DropPath. Further increasing the DropPath value did not lead to additional performance improvements.

\begin{table}[htp]
  \centering
  \caption{Ablations on Drop-path rate.}
    \renewcommand{\arraystretch}{0.3}
    \setlength\tabcolsep{50.0pt}
    \resizebox{1.0\linewidth}{!}{
    \begin{tabular}{cc}
    \toprule
    Drop-path Rate & Top-1 \\
    \midrule
    0.0     & 79.7  \\
    0.03  & 79.9 \\
    0.05  & 79.8  \\
    0.07  & 79.8  \\
    0.1   & 79.8  \\
    \bottomrule
    \end{tabular}%
    }
  \label{tab:sup_dp}%
\end{table}%

\subsection{Visualization of the ERF of different model methods}
\label{sec:supp_viserf}
\begin{figure}[t]
\centering
\includegraphics[width=0.48\textwidth]{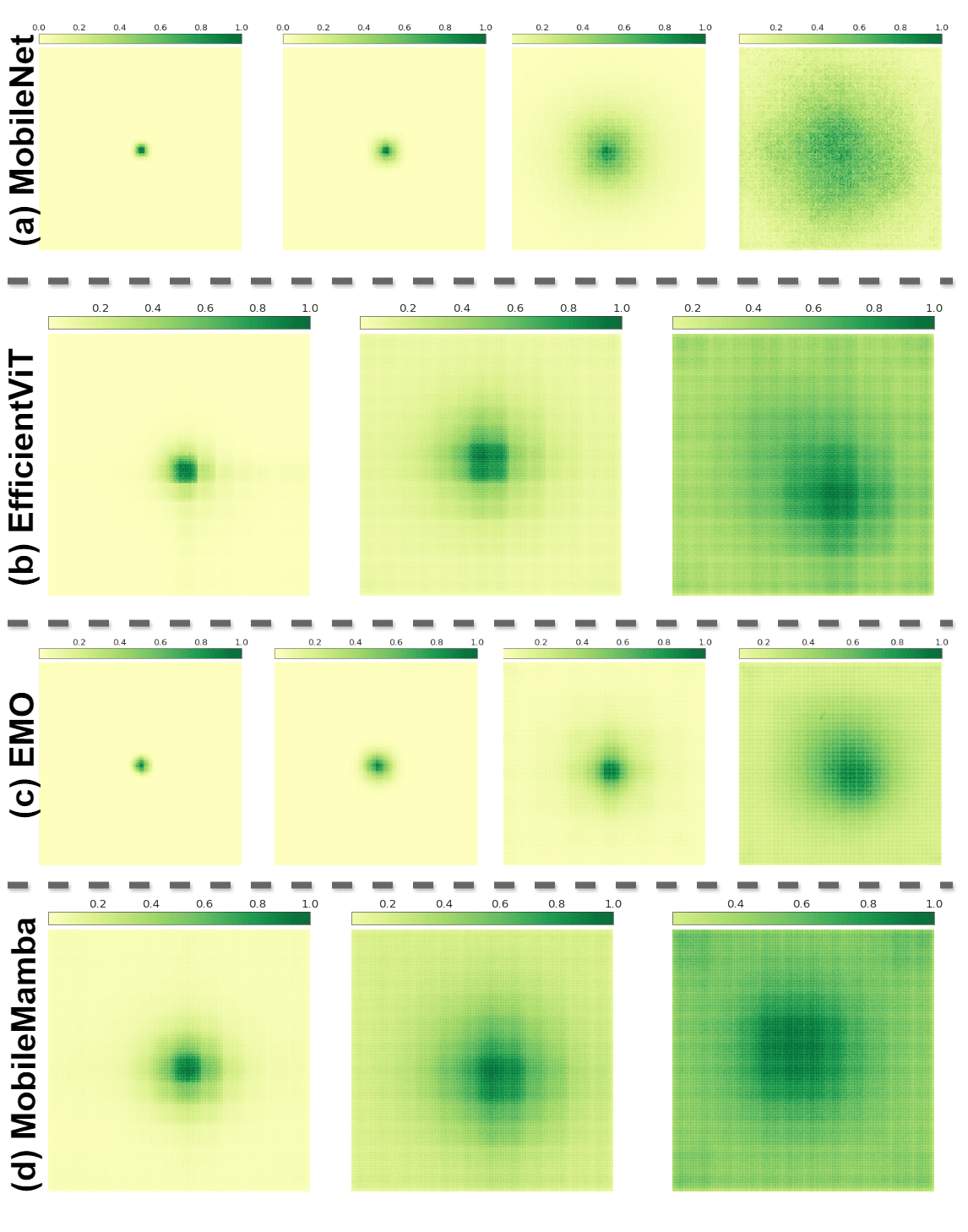} 
\caption{Visualization of the ERF of different model methods.}
\label{fig:sup_erf}
 \vspace{-0.48cm}
\end{figure}

In Fig.~\ref{fig:sup_erf}, we compare the ERF visualization results of CNN-based MobileNet~\cite{mobilenet,mobilenetv2,mobilenetv3}, Transformer-based EfficientViT~\cite{efficientvit}, hybrid-structured EMO~\cite{emo}, and our MobileMamba at different stages. The input resolution is fixed at 224x224. Both our method and EfficientViT~\cite{efficientvit} employ a three-stage approach, while MobileNet~\cite{mobilenet,mobilenetv2,mobilenetv3} and EMO~\cite{emo} follow the traditional four-stage approach. Our MobileMamba method exhibits a larger and more intense ERF at each stage compared to the other SoTAs.

\subsection{Analysis of Pre-trained Models with Different Resolutions for Downstream Tasks}
\label{sec:supp_analydown}
The specific experimental results for downstream tasks are shown in Tab.~\ref{table:supp_det_ssdlite_retinanet},\ref{table:supp_det_maskrcnn},\ref{table:supp_seg}. We investigate the impact of pre-trained model weights with different input resolutions on downstream tasks. We use two pre-trained model weights, MobileMamba-B1 and MobileMamba-B4. The only difference between them is the resolution used during pre-training on ImageNet-1K~\cite{imagenet}: MobileMamba-B1 is pre-trained at a resolution of 256, while MobileMamba-B4 is pre-trained at a resolution of 512. All other model parameters are identical.
For the object detection task in downstream tasks, MobileMamba-B1 outperforms MobileMamba-B4 on all metrics in SSDLite~\cite{ssdlite}, RetinaNet~\cite{retinanet}, and Mask RCNN~\cite{maskrcnn}. Conversely, for the semantic segmentation task, MobileMamba-B4 outperforms MobileMamba-B1 on all metrics in DeepLabv3~\cite{deeplabv3}, Semantic FPN~\cite{segfpn}, and PSPNet~\cite{pspnet}. This may be because object detection tasks require stronger semantic feature information, while semantic segmentation tasks demand higher segmentation accuracy. MobileMamba-B4, pre-trained at a high resolution of 512, extracts features with higher segmentation accuracy but slightly weaker semantic information. In contrast, MobileMamba-B1, pre-trained at a lower resolution of 256, extracts features with stronger semantic information but lower accuracy.
Therefore, we use MobileMamba-B1 pre-trained weights as the backbone for object detection tasks to enhance semantic information extraction. For semantic segmentation tasks, we use MobileMamba-B4 pre-trained weights as the backbone to improve segmentation accuracy.

\begin{table}[htp]
    \centering
    \caption{Detailed object detection performance using SSDLite~\cite{mnetv3} and RetinaNet~\cite{retinanet} of our MobileMamba on MS-COCO 2017~\cite{coco} dataset. $\dagger$: 512 $\times$ 512 resolution.}
    \label{table:supp_det_ssdlite_retinanet}
    \renewcommand{\arraystretch}{3}
    \setlength\tabcolsep{3.0pt}
    \resizebox{1.0\linewidth}{!}{
        \begin{tabular}{clcccccccc}
        \toprule[0.17em]
        & Backbone & \#Params $\downarrow$ & FLOPs $\downarrow$ & $mAP$  & $mAP^b_{50}$ & $mAP^b_{75}$ & $mAP^b_{S}$ & $mAP^b_{M}$ & $mAP^b_{L}$ \\
        \hline
        \multirow{4}{*}{\rotatebox{90}{\makecell[c]{SSDLite\\~\cite{mnetv3}}}} 
        & MobileMamba-B1 & 18.0 & 1.7G & 24.0    & 39.5  & 24.0    & 3.1   & 23.4  & 46.9  \\ 
        & MobileMamba-B4 & 18.0 & 1.7G & 23.9  & 39.5  & 24.2  & 2.9   & 23.5  & 47.1 \\ 
        \cline{2-10}
        & MobileMamba-B1$\dagger$ & 18.0 & 4.4G & 29.5  & 47.7  & 30.4  & 8.9   & 35.0    & 47.0 \\ 
        & MobileMamba-B4$\dagger$ & 18.0 & 4.4G & 29.1  & 47.1  & 30.0    & 8.7   & 34.3  & 46.7  \\ 
        \toprule[0.08em]
        \multirow{2}{*}{\rotatebox{90}{\makecell[c]{RetinaNet \\~\cite{retinanet}}}} 
        & MobileMamba-B1 & 27.1 & 151G & 39.6  & 59.8  & 42.4  & 21.5  & 43.1  & 53.9  \\ 
        & MobileMamba-B4 & 27.1 & 151G & 39.5  & 59.9  & 42.1  & 21.5  & 42.9  & 54.6 \\
        \bottomrule[0.12em]
        \end{tabular}
    }
\end{table}

\begin{table}[htp]
    \centering
    \caption{Detailed object detection performance using Mask RCNN~\cite{maskrcnn} of our MobileMamba on MS-COCO 2017~\cite{coco} dataset.}
    \label{table:supp_det_maskrcnn}
    \renewcommand{\arraystretch}{3}
    \setlength\tabcolsep{3.0pt}
    \resizebox{1.0\linewidth}{!}{
        \begin{tabular}{lcccccccc}
        \toprule[0.17em]
        \multirow{2}{*}{Backbone} & \multirow{2}{*}{\#Params $\downarrow$} & \multirow{2}{*}{FLOPs $\downarrow$} & $mAP$  & $mAP^b_{50}$ & $mAP^b_{75}$ & $mAP^b_{S}$ & $mAP^b_{M}$ & $mAP^b_{L}$ \\ 
        \cline{4-9}
        &  &  & $mAP$  & $mAP^m_{50}$ & $mAP^m_{75}$ & $mAP^m_{S}$ & $mAP^m_{M}$ & $mAP^m_{L}$ \\
        \hline
        \multirow{2}{*}{MobileMamba-B1} & \multirow{2}{*}{38.0} & \multirow{2}{*}{178G} & 40.6  & 61.8  & 43.8  & 22.4  & 43.5  & 55.9  \\ 
        \cline{4-9}
        &  &  & 37.4  & 58.9  & 39.9  & 17.1  & 39.9  & 56.4  \\ 
        \hline
        \multirow{2}{*}{MobileMamba-B4} & \multirow{2}{*}{38.0} & \multirow{2}{*}{178G} & 40.1  & 61.8  & 43.0    & 22.0    & 42.9  & 56.1   \\ 
        \cline{4-9}
        &  &  & 36.9  & 58.6  & 39.2  & 16.4  & 39.0    & 56.8  \\ 
        \toprule[0.12em]
        \end{tabular}
    }
\end{table}

\section{Detailed Downstream Results}
\subsection{Detailed Object Detection Results} \label{sec:supp_det}
Tab.~\ref{table:supp_det_ssdlite_retinanet} shows more detailed object detection results using SSDLite~\cite{mnetv3} and RetinaNet~\cite{retinanet} of our MobileMamba on MS-COCO 2017~\cite{coco} dataset, while Tab.~\ref{table:supp_det_maskrcnn} provide detailed object detection results using Mask R-CNN~\cite{maskrcnn}. 

\subsection{Detailed Semantic Segmentation Results} \label{sec:supp_seg}
Tab.~\ref{table:supp_seg} shows more detailed semantic segmentation results using DeepLabv3~\cite{deeplabv3}, Semantic FPN~\cite{segfpn}, SegFormer~\cite{segformer}, and PSPNet~\cite{pspnet} of our MobileMamba on ADE20K~\cite{ade20k} dataset.

\begin{table}[t!]
    \centering
    \caption{Detailed semantic segmentation performance using DeepLabv3~\cite{deeplabv3}, Semantic FPN~\cite{segfpn}, and PSPNet~\cite{pspnet} to adequately evaluate our MobileMamba on ADE20K~\cite{ade20k} dataset.}
    \label{table:supp_seg}
    \renewcommand{\arraystretch}{3}
    \setlength\tabcolsep{6.0pt}
    \resizebox{1.0\linewidth}{!}{
        \begin{tabular}{clccccc}
        \toprule[0.17em]
        & Backbone & \#Params $\downarrow$ & FLOPs $\downarrow$ & mIoU  & aAcc & mAcc \\
        \hline
        \multirow{2}{*}{\rotatebox{90}{\makecell[c]{DeepLabv3\\~\cite{deeplabv3}}}} 
        & MobileMamba-B1 & 23.0 & 4.7G & 36.7	& 76.0 & 46.8  \\ 
        & MobileMamba-B4 & 23.0 & 4.7G & 36.6 & 76.3 & 47.1  \\ 
        \hline
        \multirow{2}{*}{\rotatebox{90}{\makecell[c]{FPN\\~\cite{segfpn}}}} 
        & MobileMamba-B1 & 19.8 & 5.6G & 40.7 & 79.4 & 51.8 \\ 
        & MobileMamba-B4 & 19.8 & 5.6G & 42.5 & 79.9 & 53.7\\ 
        \hline
        \multirow{2}{*}{\rotatebox{90}{\makecell[c]{PSPNet\\~\cite{pspnet}}}} 
        & MobileMamba-B1 & 20.5 & 4.5G & 36.5 & 76.2 & 46.7 \\ 
        & MobileMamba-B4 & 20.5 & 4.5G & 36.9 & 76.2 & 47.9 \\ 
        \bottomrule[0.12em]
        \end{tabular}
    }
\end{table}

\end{document}